\documentclass{article}
\usepackage{iclr2027_conference,times}

\usepackage{amsmath,amsfonts,bm}

\def\eqref#1{equation~\ref{#1}}

\def\1{\bm{1}}

\DeclareMathAlphabet{\mathsfit}{\encodingdefault}{\sfdefault}{m}{sl}
\SetMathAlphabet{\mathsfit}{bold}{\encodingdefault}{\sfdefault}{bx}{n}

\usepackage{hyperref}
\usepackage{url}
\usepackage{booktabs}
\usepackage{wrapfig}
\usepackage{xcolor}
\usepackage{cleveref}
\usepackage{amssymb}
\usepackage{graphicx}
\usepackage{listings}
\usepackage{adjustbox}
\usepackage{colortbl}
\usepackage{multirow}
\usepackage{etoolbox}
\usepackage{algorithm}
\usepackage{placeins}
\usepackage{subcaption}
\usepackage{tcolorbox}

\definecolor{rowblue}{RGB}{220,240,248}

\makeatletter
\AfterEndEnvironment{algorithm}{\let\@algcomment\relax}
\AtEndEnvironment{algorithm}{\kern2pt\hrule\relax\vskip3pt\@algcomment}
\let\@algcomment\relax
\newcommand\algcomment[1]{\def\@algcomment{\footnotesize#1}}
\renewcommand\fs@ruled{\def\@fs@cfont{\bfseries}\let\@fs@capt\floatc@ruled
  \def\@fs@pre{\hrule height.8pt depth0pt \kern2pt}
  \def\@fs@post{}
  \def\@fs@mid{\kern2pt\hrule\kern2pt}
  \let\@fs@iftopcapt\iftrue}
\makeatother

\definecolor{codeblue}{RGB}{0,92,197}
\definecolor{codegreen}{RGB}{34,134,58}
\definecolor{codepurple}{RGB}{111,66,193}

\title{PixelMoE: Scaling Pixel-Space Diffusion with LLM-Style Mixture-of-Experts}

\author{\textbf{Ruixiao Dong}$^{1,2}\thanks{This work was conducted during the author's internship at KlingAI.}$
\quad
\textbf{Jianmin Bao}$^{2}\thanks{Corresponding authors.}$
\quad
\textbf{Bozhi Luan}$^{1,2}\footnotemark[1]$
\quad
\textbf{Zijie Meng}$^{2,3}\footnotemark[1]$
\quad
\textbf{Xin Tao}$^{2}$
\quad
\textbf{Pengfei Wan}$^{2}$
\\
\textbf{Kun Gai}$^{2}$
\quad
\textbf{Li Li}$^{1}\footnotemark[2]$
\quad
\textbf{Houqiang Li}$^{1}$
\\
$^{\text{1}}$ University of Science and Technology of China \\
$^{\text{2}}$ KlingAI Research \\
$^{\text{3}}$ Peking University \\
{\tt\small dongruixiaoyx@mail.ustc.edu.cn baojianmin@kuaishou.com} \\
{\tt\small \{lil1,lihq\}@ustc.edu.cn} \\
}

\iclrfinalcopy
\begin{document}
\maketitle

\vspace{-0.2cm}
\begin{abstract}
Pixel-space diffusion models provide an appealing alternative to latent-space diffusion models by directly modeling raw images without relying on a learned tokenizer.
However, scaling pixel-space models has predominantly followed dense strategies, tightly coupling model capacity with active computation.
In this work, we show that standard LLM-style Mixture-of-Experts (MoE) architectures can be effectively transferred to pixel-space diffusion.
Building on this finding, we introduce \textbf{PixelMoE}, a pixel-space generative framework that sparsely scales model capacity with standard LLM-style MoE, without diffusion-specific routing mechanisms or auxiliary routing objectives.
Specifically, PixelMoE combines standard Token-Choice (TC) routing with fine-grained routed experts for input-adaptive sparse computation, a shared expert for capturing common representations, and a bias-based load-balancing strategy.
To complement sparse capacity scaling, PixelMoE further introduces Pixel-Space Hierarchical Flow Matching, which varies spatial resolution along the denoising trajectory from global structure formation to fine-grained detail refinement.
At comparable active parameter counts, PixelMoE consistently outperforms its dense counterparts across model scales, demonstrating effective sparse capacity scaling in pixel space.
On class-conditional ImageNet-1K, PixelMoE achieves FID scores of 1.47 at $256\times256$ and 1.53 at $512\times512$ without representation alignment or perceptual losses, comparing favorably with leading pixel-space and latent-space generative models.
Code and pretrained models are available at \url{https://github.com/KlingAIResearch/PixelMoE} and \url{https://huggingface.co/KlingTeam/PixelMoE}, respectively.
\end{abstract}

\begin{figure*}[h]
    \centering
    \vspace{-0.2cm}
    \includegraphics[width=0.94\linewidth]{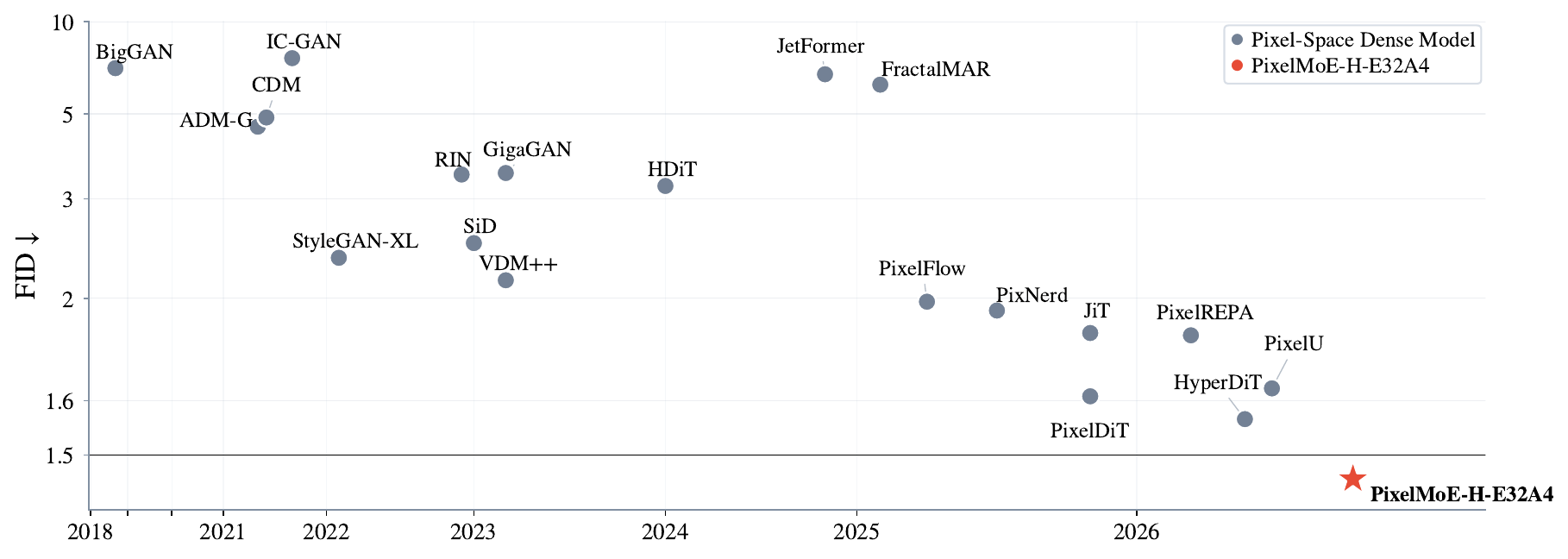}
    \caption{
    \textbf{Progress in pixel-space image generation on ImageNet-1K at $256\times256$.}
    PixelMoE sparsely scales model capacity with MoE and compares favorably with leading approaches.
    }
    \label{fig:pixel_evolution}
\end{figure*}

\section{Introduction}
Recent advances in diffusion models, particularly Diffusion Transformers (DiTs)~\citep{dit,ma2024sit} trained with flow matching objectives~\citep{lipman2023flow,liu2023flow}, have substantially improved the quality and scalability of image generation.
Most modern diffusion models~\citep{rombach2022high} denoise in latent spaces extracted by variational autoencoders (VAEs)~\citep{kingma2013auto}, whose lossy compression may discard high-frequency details and introduce structural artifacts, leading to a reconstruction bottleneck that limits achievable image fidelity.
Recently, pixel-space diffusion models~\citep{dhariwal2021diffusion,li2026back} have regained attention by directly modeling raw images rather than learned latent representations.
Advances in clean-image prediction~\citep{li2026back}, architectural design~\citep{he2026hyperdit,yu2026pixeldit}, and training objectives~\citep{ma2026pixelgen} have made raw-pixel modeling easier, enabling pixel-space models to achieve performance competitive with leading latent-space approaches.

\begin{figure*}[t]
    \centering
    \includegraphics[width=\linewidth]{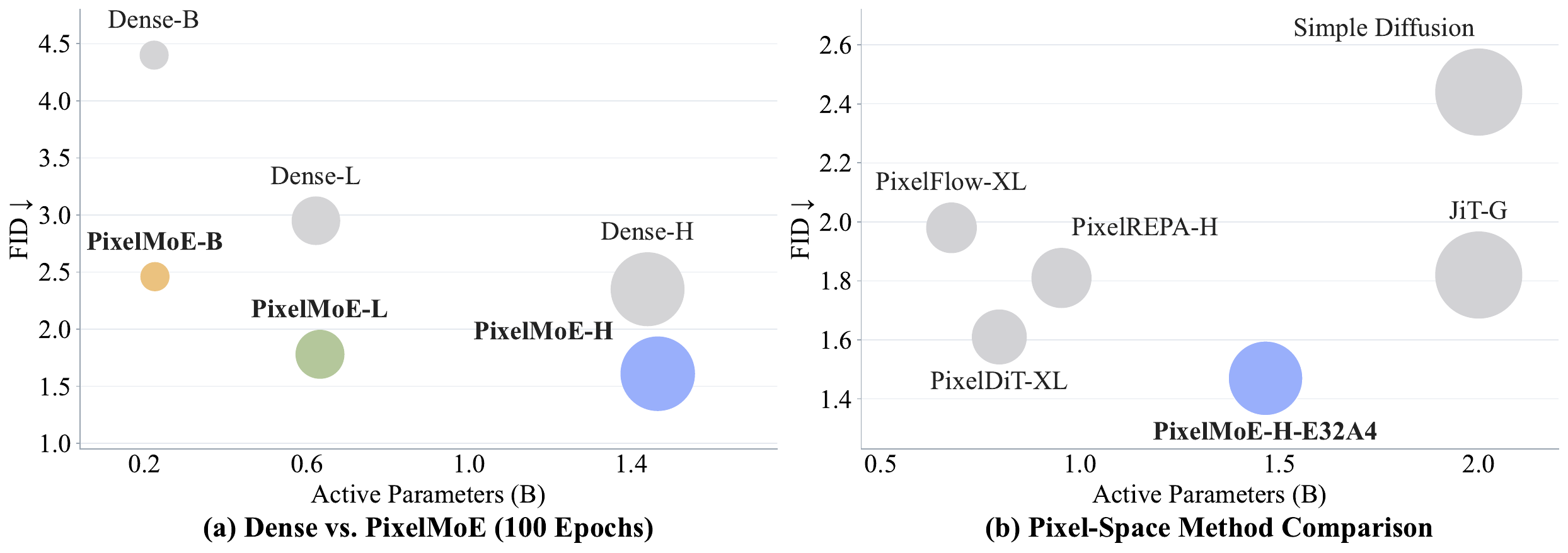}
    \caption{
    \textbf{ImageNet-1K generation with PixelMoE.}
      Left: Dense \textit{vs.} PixelMoE after 100 training epochs, where PixelMoE consistently achieves lower FID at comparable active parameter counts.
      PixelMoE-B/L/H abbreviate PixelMoE-B/L/H-E32A4, respectively, with 32 routed experts and top-4 activation.
      Right: Comparison with representative pixel-space methods.
      Marker size denotes active parameter counts.
    }
    \label{fig:performance}
\end{figure*}

Despite this progress, sparsely scaling pixel-space diffusion remains largely unexplored.
Existing pixel-space models primarily scale their generators densely by increasing model width and depth, tightly coupling capacity growth with active computation.
In contrast, Mixture-of-Experts (MoE) architectures~\citep{shazeer2017} expand total capacity through sparse expert activation while controlling active computation, and have become a standard scaling paradigm in Large Language Models (LLMs)~\citep{liu2024deepseek}.
However, directly transferring standard LLM-style MoE to diffusion has not consistently yielded clear gains over dense counterparts~\citep{fei2024scaling,liu2025efficient}.
Subsequent approaches therefore adopt alternative routing paradigms such as Expert-Choice (EC)~\citep{sun2025ec,shi2025diffmoe}, or introduce diffusion-specific routing guidance and objectives~\citep{wei2026routing}.
This naturally raises the question:
\begin{tcolorbox}[
    colback=gray!12,
    colframe=gray!12,
    boxrule=0pt,
    arc=2mm,
    boxsep=0pt,
    left=8pt,
    right=8pt,
    top=6pt,
    bottom=6pt
]
    \centering
    \textbf{\textit{
        Can standard LLM-style MoE be directly transferred to pixel-space diffusion?
    }}
\end{tcolorbox}

We answer this question affirmatively.
We propose \textbf{PixelMoE}, a \textbf{Pixel-space Mixture-of-Experts} generative framework that sparsely scales pixel-space diffusion transformers without diffusion-specific routing mechanisms or auxiliary routing objectives.
Specifically, PixelMoE adopts standard Token-Choice (TC) routing, where each image token dynamically selects its top-$A$ experts based on the routing scores.
Each MoE layer combines fine-grained routed experts for input-adaptive sparse computation with a shared expert that captures common representations across image tokens.
To maintain balanced expert utilization without auxiliary balancing objectives, we further adopt a bias-based load-balancing strategy.
Beyond sparse capacity scaling, pixel-space representations can be directly downsampled and upsampled across resolutions while preserving meaningful visual structure.
We exploit this property with Pixel-Space Hierarchical Flow Matching, organizing generation into a low-resolution coarse stage followed by full-resolution refinement.

Extensive experiments demonstrate strong sparse scaling behavior of PixelMoE.
Under comparable active parameter counts, PixelMoE outperforms its dense counterparts across model scales.
Furthermore, increasing the number of routed experts consistently improves generation quality.
Standard Token-Choice routing also substantially outperforms Expert-Choice routing, without diffusion-specific routing guidance.
Moreover, our routing analysis reveals clear expert specialization across semantic classes and along the denoising trajectory without explicit semantic or timestep guidance.
Together, these results suggest that effective sparse scaling does not require diffusion-specific routing: with appropriately structured sparse capacity, standard Token-Choice routing can effectively exploit the rich semantic and denoising information already encoded in pixel-space representations.
On ImageNet-1K~\citep{deng2009imagenet}, PixelMoE achieves FID scores of 1.47 at $256\times256$ and 1.53 at $512\times512$ without representation alignment~\citep{yu2025representation,shin2026representation} or perceptual losses~\citep{ma2026pixelgen}, comparing favorably with leading pixel-space and latent-space generative models.
These results demonstrate that LLM-style sparse scaling transfers effectively to pixel-space diffusion.

\section{Related Work}
\paragraph{Pixel-Space Generative Models.}
Early generative models, including GANs~\citep{brock2018large,casanova2021instance} and diffusion models~\citep{dhariwal2021diffusion,ho2022cascaded}, synthesize images directly in pixel space.
Latent-space models~\citep{rombach2022high,esser2021taming} later became dominant by modeling compact representations extracted by variational autoencoders (VAEs)~\citep{kingma2013auto}, whose lossy compression introduces a reconstruction bottleneck that limits the achievable generation fidelity.
This limitation has renewed interest in pixel-space generative models, which directly model raw images without lossy compression.
Simple Diffusion~\citep{hoogeboom2023simple} demonstrated competitive high-resolution pixel-space generation through architectural scaling, noise schedules, and spatial downsampling, while VDM++~\citep{kingma2023understanding} studied diffusion-objective weighting and noise augmentation.
JiT~\citep{li2026back} further showed that a large-patch Transformer can effectively model raw pixels through clean-image prediction.
Subsequent approaches improve pixel-space generation through multi-resolution generation~\citep{chen2025pixelflow}, neural-field decoding~\citep{wang2026pixnerd}, hierarchical architectures~\citep{yu2026pixeldit,he2026hyperdit,guo2026pixelu}, frequency-decoupled modeling~\citep{ma2026deco}, representation alignment~\citep{shin2026representation}, perceptual supervision~\citep{ma2026pixelgen}, and alternative flow parameterizations~\citep{chen2026asymmetric}.
Despite these advances, most existing pixel-space models remain dense, coupling increased model capacity with increased active computation, while sparse MoE scaling in pixel-space diffusion remains underexplored.

\paragraph{Mixture-of-Experts in Visual Generation.}
The Mixture-of-Experts (MoE) paradigm was introduced to enable specialization among multiple expert networks~\citep{jacobs1991adaptive} and later extended with sparse gating~\citep{shazeer2017}.
Modern MoE architectures~\citep{fedus2022switch,dai2024deepseekmoe} have become a key scaling mechanism for Large Language Models (LLMs)~\citep{yang2025qwen3,liu2024deepseek}, motivating their adoption in visual generation.
Early expert-based diffusion models, largely built on U-Net backbones, partition the denoising trajectory by noise level, denoising task, or generation path and assign specialized parameters to different regimes~\citep{balaji2022ediff,feng2023ernie,xue2023raphael,lee2024multi,ganjdanesh2024mixture}.
As diffusion backbones shift toward Diffusion Transformers (DiTs)~\citep{dit}, recent work instead explores sparse expert routing within Transformer blocks.
However, directly applying conventional Token-Choice routing has not consistently yielded clear gains over dense models~\citep{fei2024scaling}.
Subsequent approaches have therefore introduced Expert-Choice routing~\citep{sun2025ec}, diffusion-aware routing or dynamic capacity allocation~\citep{cheng2025diff,shi2025diffmoe}, and specialized token-expert assignment with auxiliary routing objectives~\citep{yuan2025expert,wei2026routing}.
However, most existing sparse DiT approaches focus on latent-space diffusion.
JiTMoE~\citep{liu2025efficient} has recently extended MoE to pixel-space diffusion but reports only modest gains over its dense JiT counterparts.
We therefore investigate whether standard LLM-style sparse MoE can provide systematic scaling gains in pixel-space generation.

\section{PixelMoE: Pixel-Space Diffusion with Mixture-of-Experts}

\begin{figure*}[t]
    \centering
    \includegraphics[width=\linewidth]{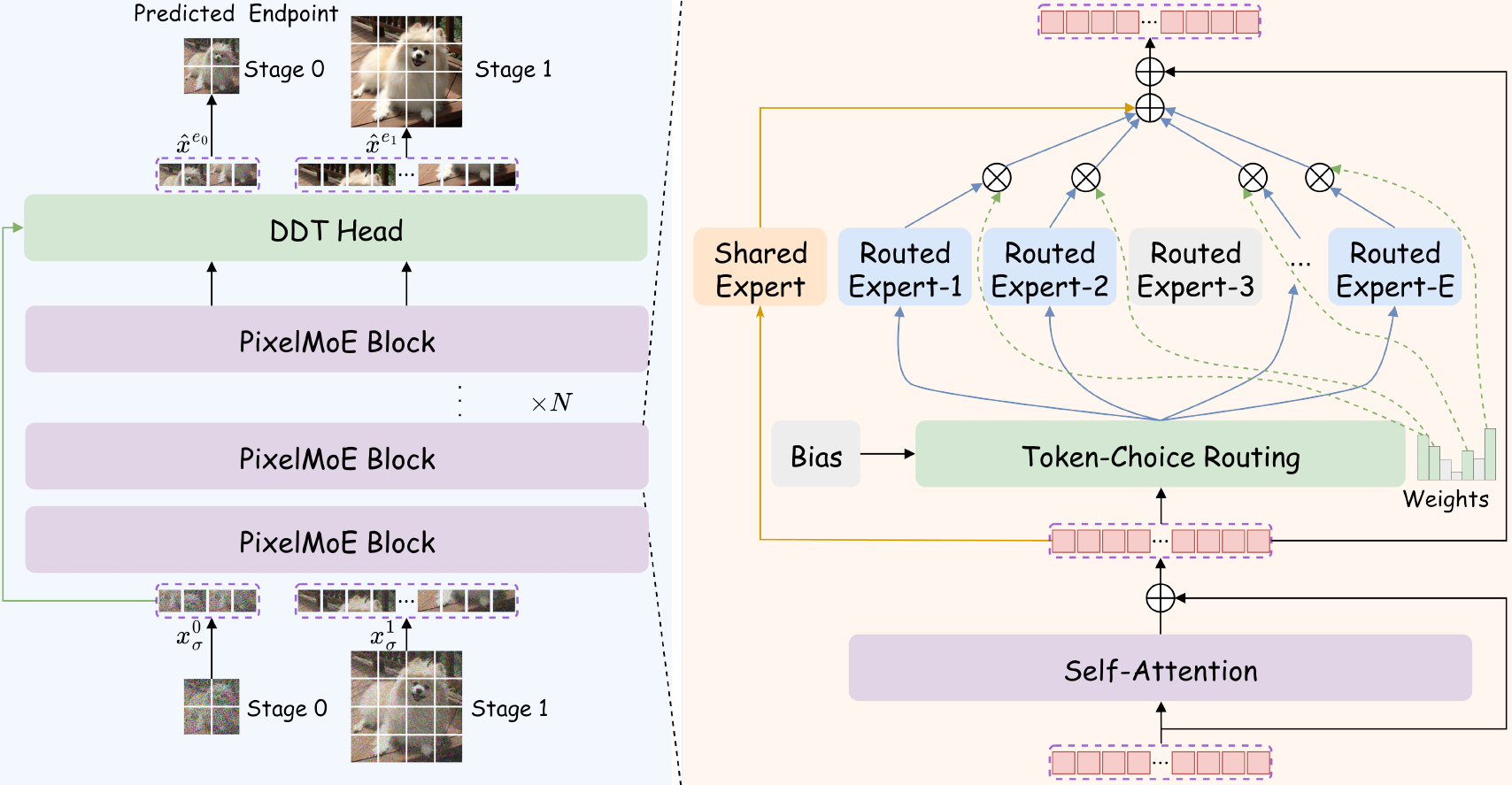}
    \caption{
   \textbf{Overview of PixelMoE.}
    Left: PixelMoE performs Pixel-Space Hierarchical Flow Matching using a shared PixelMoE backbone and DDT prediction head; the initial dense blocks are omitted for clarity.
    Right: Each PixelMoE block consists of self-attention followed by an LLM-style sparse MoE layer with Token-Choice routed experts and a shared expert.
    A balancing bias is employed to maintain balanced expert utilization.
    RMSNorm and timestep/class conditioning are omitted.
    }
    \label{fig:framework}
\end{figure*}

PixelMoE is a pixel-space generative framework designed to sparsely scale model capacity with standard LLM-style Mixture-of-Experts (MoE), while employing Pixel-Space Hierarchical Flow Matching for coarse-to-fine generation.
As illustrated in the left panel of~\Cref{fig:framework}, generation proceeds through two stages at different spatial resolutions.
Stage 0 performs partial denoising at low resolution to establish global structure, while Stage 1 continues denoising at full resolution for detail refinement.
The two stages share the PixelMoE backbone and Decoupled Diffusion Transformer (DDT)~\citep{wang2026ddt} prediction head.
The backbone consists of initial dense blocks followed by sparse PixelMoE blocks, while the DDT head performs stage-wise endpoint prediction.

As shown in the right panel of~\Cref{fig:framework}, each PixelMoE block consists of a self-attention layer followed by an LLM-style sparse MoE layer.
For each image token, standard Token-Choice routing selects a subset of fine-grained routed experts, while a shared expert is always activated to capture common representations across image tokens.
To maintain balanced expert utilization without an auxiliary load-balancing objective, we further introduce a balancing bias during routing.
Importantly, the router operates directly on pixel-token representations without explicit semantic or timestep routing guidance.
We detail the sparse MoE design and the hierarchical flow formulation in the following subsections.

\subsection{LLM-Style Mixture-of-Experts for Pixel-Space Diffusion}
\label{sec:moe}
The core design of PixelMoE is to directly transfer the standard LLM-style MoE paradigm to pixel-space diffusion.
Following modern LLM-style MoE architectures~\citep{liu2024deepseek,team2026kimi}, we retain dense feed-forward networks (FFNs) in the early Transformer blocks and replace the FFNs in the remaining blocks with sparse MoE layers.
Each MoE layer adopts standard Token-Choice routing over pixel-token representations, combines fine-grained routed experts with a shared expert, and employs a bias-based load-balancing strategy without an auxiliary objective.

\paragraph{Token-Choice Routing.}
For each image token, Token-Choice (TC) routing first computes its affinity to all routed experts.
Let $\mathbf{h}^{s}_{t,i}\in\mathbb{R}^{d}$ denote the hidden state of image token $i$ at global timestep $t$ in stage $s$, and let $E$ denote the number of routed experts.
We use a bias-free linear router with weight $\mathbf{W}_{r}\in\mathbb{R}^{E\times d}$ to produce sigmoid routing scores:
\begin{equation}
\mathbf{r}^{s}_{t,i}=\operatorname{sigmoid}\left(\mathbf{W}_{r}\mathbf{h}^{s}_{t,i}\right),
\end{equation}
where $r^{s}_{t,i,e}$ denotes the routing score of token $i$ for the $e$-th routed expert.
To regulate expert utilization, each routed expert is associated with a balancing bias $b_e$, which is added to the routing score only during expert selection.
The set of top-$A$ activated experts is
\begin{equation}
\mathcal{S}^{s}_{t,i}=\operatorname{TopA}_{e}\left(r^{s}_{t,i,e}+b_e\right),
\end{equation}
where $A$ denotes the number of routed experts activated for each image token.
After selection, the balancing bias is no longer used and the routing scores are normalized over the selected experts:
\begin{equation}
g^{s}_{t,i,e}=\frac{r^{s}_{t,i,e}}{\sum_{j\in\mathcal{S}^{s}_{t,i}}r^{s}_{t,i,j}},\qquad e\in\mathcal{S}^{s}_{t,i},
\end{equation}
where $g^{s}_{t,i,e}$ denotes the routing weight of expert $e$ for token $i$.
Notably, the router relies solely on image-token hidden states, without explicit semantic or timestep guidance.

\paragraph{Expert Aggregation.}
Given the selected experts $\mathcal{S}^{s}_{t,i}$ and routing weights $g^{s}_{t,i,e}$, PixelMoE aggregates the routed-expert outputs with the output of an always-active shared expert.
Let $\mathcal{E}_e(\cdot)$ denote the $e$-th routed expert and $\mathcal{E}_{\mathrm{sh}}(\cdot)$ the shared expert.
Only the experts in $\mathcal{S}^{s}_{t,i}$ process the image-token hidden state $\mathbf{h}^{s}_{t,i}$ in the routed branch, while the shared expert is applied to every image token.
Their outputs are combined as
\begin{equation}
\mathbf{y}^{s}_{t,i}
=
\lambda_r
\sum_{e\in\mathcal{S}^{s}_{t,i}}
g^{s}_{t,i,e}\,
\mathcal{E}_e\!\left(\mathbf{h}^{s}_{t,i}\right)
+
\mathcal{E}_{\mathrm{sh}}\!\left(\mathbf{h}^{s}_{t,i}\right),
\end{equation}
where $\lambda_r=\sqrt{A}$ denotes the routing scale.
Both routed and shared experts are implemented as FFNs with reduced intermediate dimensions, such that their combined active parameter count is comparable to that of the dense baseline.

\paragraph{Bias-Based Load Balancing.}
To maintain balanced expert utilization without introducing an auxiliary load-balancing objective, we dynamically update the expert-wise balancing biases during training.
We collect the balancing biases for the routed experts into $\mathbf{b}=(b_1,\ldots,b_E)\in\mathbb{R}^{E}$.
The expert-load distribution in the current training batch is $\mathbf{F}\in\mathbb{R}^{E}$, obtained by averaging image-token assignments, with each selected expert contributing $1/A$, while the uniform target is $\mathbf{Q}=(1/E,\ldots,1/E)$.
We update the balancing biases as
\begin{equation}
\mathbf{b}
\leftarrow
\mathbf{b}
-
\gamma
\frac{\mathbf{F}-\mathbf{Q}}
{\operatorname{RMS}(\mathbf{F}-\mathbf{Q})},
\end{equation}
where $\gamma=0.001$ controls the update magnitude.
The detailed computation of $\mathbf{F}$ and the RMS normalization is provided in~\Cref{app:bias_balancing}.

\subsection{Pixel-Space Hierarchical Flow Matching}
\label{sec:hierarchical}
\begin{figure*}[t]
    \centering
    \includegraphics[width=1.0\linewidth]{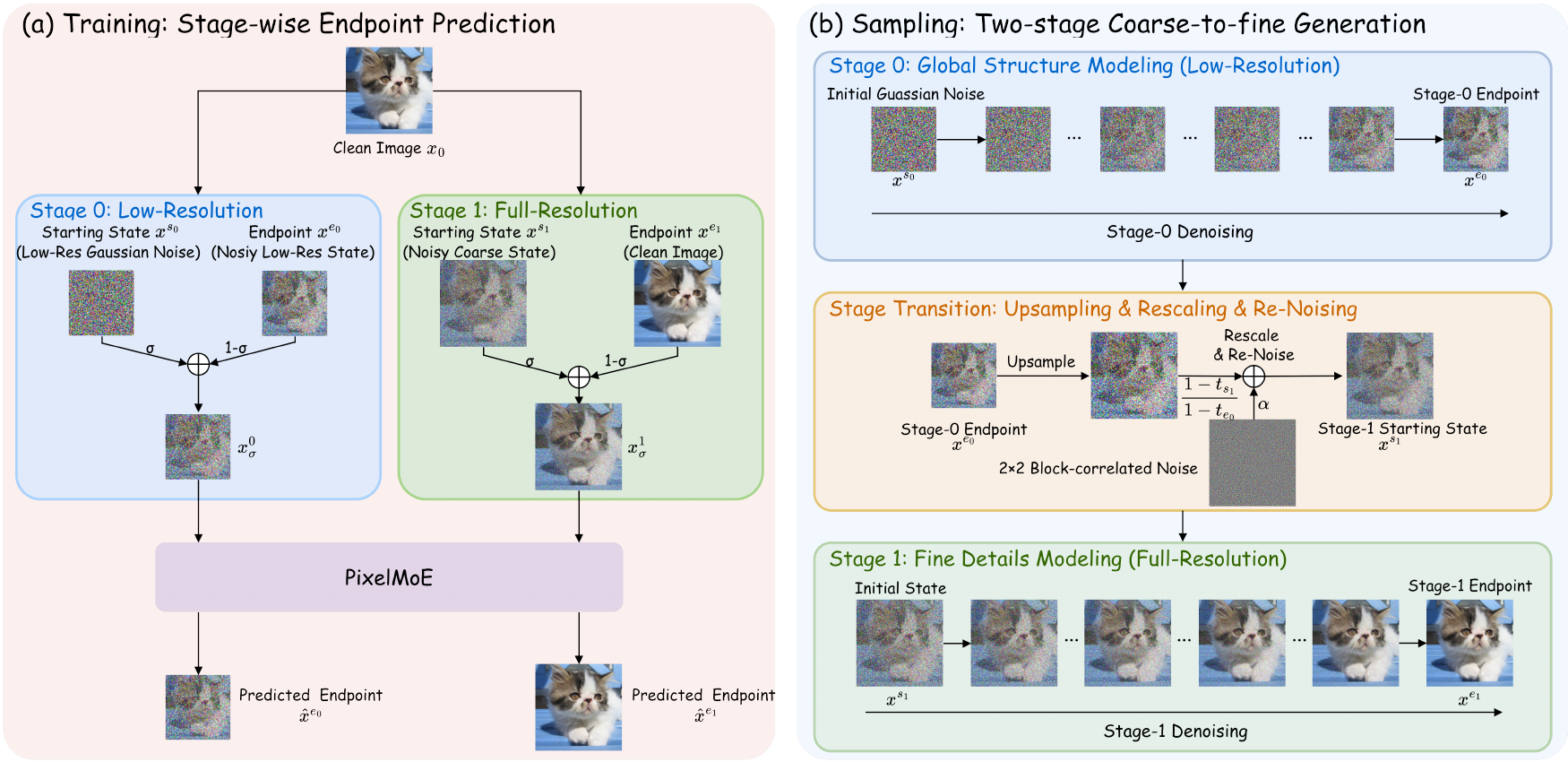}
    \caption{
    \textbf{Pixel-Space Hierarchical Flow Matching in PixelMoE.} 
    PixelMoE is jointly trained on a low-resolution coarse stage and a full-resolution refinement stage with stage-wise endpoint prediction.
    During sampling, generation first follows the low-resolution Stage 0 flow to obtain a coarse endpoint, which is then transitioned to the Stage 1 starting state for full-resolution refinement.
    }
    \label{fig:hierarchical_flow}
\end{figure*}

Pixel space naturally supports coarse-to-fine generation, as spatial resampling directly produces meaningful multi-resolution images.
We therefore vary the spatial resolution along the denoising trajectory, using coarse representations during early denoising and full-resolution representations for later refinement.
This hierarchical trajectory also requires the model to process heterogeneous pixel-token representations across resolutions and noise levels, which are naturally accommodated by the input-adaptive routing of PixelMoE.
Specifically, Pixel-Space Hierarchical Flow Matching organizes generation into two stages: Stage 0 performs partial denoising at low resolution for coarse structure modeling, while Stage 1 continues at full resolution for detail refinement.
As illustrated in~\Cref{fig:hierarchical_flow}, both stages share the same PixelMoE model while denoising at different spatial resolutions and noise levels.

\paragraph{Hierarchical Flow Trajectory.}
Given a clean image $x_0$, Stage 0 performs partial denoising at low resolution, starting from Gaussian noise $x^{s_0}$ and terminating at a noisy coarse endpoint $x^{e_0}$:
\begin{equation}
\begin{aligned}
    x^{s_0} &= \epsilon_0, \qquad
    \epsilon_0\sim\mathcal{N}(0,\delta_n^2 I),\\
    x^{e_0} &= (1-t_{e_0})\mathcal{D}(x_0)+t_{e_0}\epsilon_0,
\end{aligned}
\end{equation}
where $\mathcal{D}(\cdot)$ denotes $2\times$ bilinear downsampling and $t_{e_0}$ specifies the Stage 0 endpoint.
We use $\delta_n=1$ for $256\times256$ generation and $\delta_n=2$ for $512\times512$ generation.

Stage 1 then continues denoising at full resolution from a noisy coarse state:
\begin{equation}
\begin{aligned}
    x^{s_1} &= (1-t_{s_1})\widetilde{x}_0+t_{s_1}\epsilon_1, \qquad
    \epsilon_1\sim\mathcal{N}(0,\delta_n^2 I),\\
    x^{e_1} &= x_0,
\end{aligned}
\end{equation}
where $\widetilde{x}_0=\mathcal{U}(\mathcal{D}(x_0))$ is the coarse full-resolution image obtained by downsampling and upsampling $x_0$, $\mathcal{U}(\cdot)$ denotes $2\times$ nearest-neighbor upsampling, and $t_{s_1}$ specifies the Stage 1 starting noise level.
Accordingly, $t_{s_0}=1$ and $t_{e_1}=0$ correspond to the pure-noise start of Stage 0 and the clean-image endpoint of Stage 1, respectively.
Within each stage $s\in\{0,1\}$, we introduce a stage-relative timestep $\sigma\in[0,1]$ and linearly interpolate between the stage start and endpoint:
\begin{equation}
    x_{\sigma}^{s}=(1-\sigma)x^{e_s}+ \sigma x^{s_s},
\end{equation}
where $x^{s_s}$ and $x^{e_s}$ denote the start and endpoint of stage $s$, respectively; \textit{e.g.}, $x^{s_0}$ and $x^{e_0}$ for Stage 0.
The corresponding target velocity is
\begin{equation}
    v^{s}=\frac{\partial x_{\sigma}^{s}}{\partial \sigma}=x^{s_s}-x^{e_s}.
\end{equation}

\paragraph{Stage-Wise Endpoint Prediction.}
Inspired by the clean-data prediction in JiT~\citep{li2026back}, we predict the endpoint of each stage.
Specifically, given $x_{\sigma}^{s}$, PixelMoE predicts the stage endpoint $\hat{x}_{\theta}^{e_s}(x_{\sigma}^{s},t,c)$, where $t=t_{e_s}+\sigma(t_{s_s}-t_{e_s})$ is the corresponding global timestep used for conditioning the model and $c$ denotes the class condition.
The predicted endpoint induces the velocity
\begin{equation}
   v_{\theta}(x_{\sigma}^{s},t,c)=\frac{(x_{\sigma}^{s}-\hat{x}_{\theta}^{e_s}(x_{\sigma}^{s},t,c))}{\sigma}. 
\end{equation}
Substituting this parameterization into the velocity-matching objective gives
\begin{equation}
\mathcal{L}_{\mathrm{v\text{-}endpoint}}=\mathbb{E}\left[\mathcal{L}_{\mathrm{MSE}}\!\left(v_{\theta}(x_{\sigma}^{s},t,c),v^{s}\right)\right]=\mathbb{E}\left[\frac{1}{\sigma^2}\mathcal{L}_{\mathrm{MSE}}\!\left(\hat{x}_{\theta}^{e_s}(x_{\sigma}^{s},t,c),x^{e_s}\right)\right],
\end{equation}
where $\mathcal{L}_{\mathrm{MSE}}$ denotes the mean squared error. 
For numerical stability, we clamp $\sigma$ to $\max(\sigma,\sigma_{\min})$ in the endpoint-prediction loss during training and in the velocity computation during sampling, with $\sigma_{\min}=0.05$.
This velocity-space endpoint objective is the only generation loss used to train the default PixelMoE models; we do not employ representation alignment or perceptual losses.

\paragraph{Coarse-to-Fine Sampling.}
During sampling, generation proceeds sequentially from Stage 0 to Stage 1.
Starting from low-resolution Gaussian noise, PixelMoE first follows the Stage 0 flow to obtain the coarse endpoint $\hat{x}_{\theta}^{e_0}$.
The upsampled Stage 0 endpoint cannot be directly used as the Stage 1 starting state $\hat{x}_{\theta}^{s_1}$, since both its signal coefficient and noise covariance differ from those of the predefined Stage 1 starting distribution.
We therefore rescale the upsampled endpoint and inject an additional Gaussian perturbation:
\begin{equation}
    \hat{x}_{\theta}^{s_1}=\frac{1-t_{s_1}}{1-t_{e_0}}\mathcal{U}\!\left(\hat{x}_{\theta}^{e_0}\right)+\alpha\epsilon',
\end{equation}
where the scaling factor aligns the signal component with the Stage 1 starting state, while $\epsilon'$ follows a block-correlated Gaussian distribution and $\alpha$ is chosen to match the target covariance.
The resulting state is constructed to match the predefined Stage 1 statistics while preserving the coarse structure generated in Stage 0.
PixelMoE then follows the Stage 1 flow toward the clean full-resolution endpoint.
The transition coefficients and block-correlated noise construction are derived in~\Cref{app:hierarchical_transition}.

\paragraph{Stage-Aware Timestep Sampling.}
The two stages denoise at different spatial resolutions and cover different portions of the denoising trajectory.
We therefore adopt a stage-aware timestep sampling strategy that separately controls stage allocation and stage-relative timestep sampling.
Stage 0 covers the global timestep interval $[t_{e_0},1]$, while Stage 1 covers $[0,t_{s_1}]$.
Specifically, a global $\operatorname{LogitNormal}(\mu_{\mathrm g},\delta_{\mathrm g})$ distribution determines the sampling probability of each stage, with the probability mass in the overlapping region divided equally between them.
After sampling a stage $s$, we independently sample the stage-relative timestep $\sigma\sim\operatorname{LogitNormal}(\mu_{\mathrm l},\delta_{\mathrm l})$ and map it to the corresponding global timestep as
\begin{equation}
  t=t_{e_s}+\sigma(t_{s_s}-t_{e_s}). 
\end{equation}
This formulation decouples stage allocation from stage-relative timestep sampling, allowing the timestep distribution within each stage to be controlled independently of how frequently that stage is selected.

\section{Experiments}

\begin{figure*}[t]
    \centering
    \begin{minipage}[t]{0.46\textwidth}
        \vspace{0pt}
        \centering
        \makeatletter\def\@captype{table}\makeatother
        \setlength{\tabcolsep}{5.5pt}
        \renewcommand{\arraystretch}{1.05}
        \resizebox{\linewidth}{!}{%
        \begin{tabular}{lccc}
            \toprule
            Method & \#Params & \#Active & FID$\downarrow$ \\
            \midrule
            Dense-B & 223M & 223M & 4.40 \\
            \rowcolor{rowblue}
            \textbf{PixelMoE-B-E32A4} & 461M & 225M & \textbf{2.46} \\
            \midrule
            Dense-L & 622M & 622M & 2.95 \\
            \rowcolor{rowblue}
            \textbf{PixelMoE-L-E32A4} & 1.65B & 632M & \textbf{1.78} \\
            \midrule
            Dense-H & 1.44B & 1.44B & 2.35 \\
            \rowcolor{rowblue}
            \textbf{PixelMoE-H-E32A4} & 3.68B & 1.47B & \textbf{1.61} \\
            \bottomrule
        \end{tabular}%
        }
        \caption{\textbf{Active parameter scaling: dense \textit{vs.} MoE.}}
        \label{tab:activate_scaling}
    \end{minipage}%
    \hfill
    \begin{minipage}[t]{0.46\textwidth}
        \vspace{0pt}
        \centering
        \makeatletter\def\@captype{figure}\makeatother
        \includegraphics[width=\linewidth]{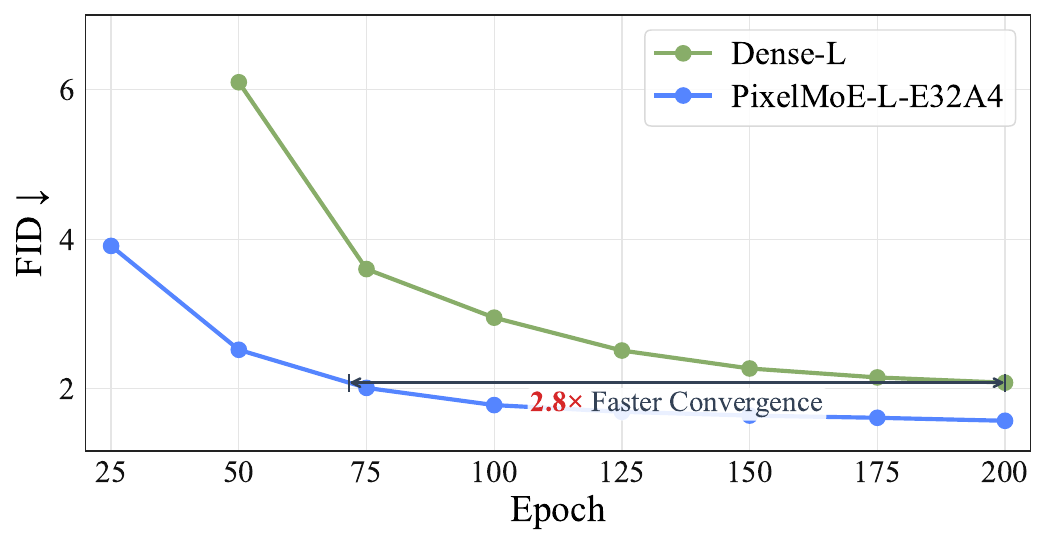}
        \vspace{-0.5cm}
        \caption{\textbf{Convergence comparison between Dense-L and PixelMoE-L-E32A4.}}
        \label{fig:epoch_fid_large}
    \end{minipage}
\end{figure*}

We conduct all experiments on class-conditional ImageNet-1K generation~\citep{deng2009imagenet}.

\paragraph{Training Details.}
During training, we apply random horizontal flipping with probability $0.5$.
Models for comparison with previous methods are trained for $320$ epochs, while other experiments use $100$ epochs unless otherwise specified.
For PixelMoE-H-E32A4, we apply additional regularization after epoch $200$.
We use a learning rate of $2\times10^{-4}$ with a $5$-epoch linear warm-up, followed by a constant schedule with cosine decay over the final $40$ epochs.
For optimization, we apply Gram Newton-Schulz Muon (gMuon)~\citep{jordan2024muon,GramNewtonSchulz} with momentum $0.95$ to eligible matrix-valued parameters, while the remaining parameters are optimized with AdamW~\citep{loshchilov2018decoupled} using $(\beta_1,\beta_2)=(0.9,0.95)$.
We employ stage-aware Logit-Normal timestep sampling with
$(\mu_{\mathrm g},\delta_{\mathrm g})=(0.8,0.8)$ for stage allocation and
$(\mu_{\mathrm l},\delta_{\mathrm l})=(0.6,0.8)$ for stage-relative timestep sampling.
Two exponential moving average (EMA) copies with decay rates $0.9996$ and $0.9999$ are maintained during training.
Detailed training configurations are provided in~\Cref{app:training_details}.

\paragraph{Evaluation Protocol.}
We evaluate generation quality using Fréchet Inception Distance (FID)~\citep{heusel2017gans}, Inception Score (IS)~\citep{salimans2016improved}, Precision, and Recall~\citep{kynkaanniemi2019improved}.
We sample 50,000 images with labels uniformly distributed over the ImageNet-1K classes.
We use a 50-point global timestep grid with the Heun solver and apply classifier-free guidance (CFG)~\citep{ho2021classifierfree} with a CFG interval~\citep{kynkaanniemi2024applying}.
Detailed evaluation settings are provided in~\Cref{app:evaluation_details}.

\subsection{Sparse Scaling Analysis}
We first investigate whether sparse MoE scaling can effectively improve pixel-space generation.
We study the sparse scaling behavior of PixelMoE from three perspectives: active parameter scaling, expert scaling, and computation scaling via token granularity.

\begin{table*}[t]
\centering
\small
\begin{subtable}[t]{0.44\textwidth}
    \vspace{0pt}
    \centering
    \begin{minipage}[t][2.6cm][t]{\linewidth}
        \vspace{0pt}
        \centering
        \setlength{\tabcolsep}{3.5pt}
        \resizebox{\linewidth}{!}{%
        \begin{tabular}{lccc}
            \toprule
            Method & \#Params & \#Active & FID$\downarrow$ \\
            \midrule
            PixelMoE-L-E8A4 & 892M & 632M & 2.28 \\
            PixelMoE-L-E16A4 & 1.14B & 632M & 2.03 \\
            \rowcolor{rowblue}
            \textbf{PixelMoE-L-E32A4} & 1.65B & 632M & \textbf{1.78} \\
            \bottomrule
        \end{tabular}%
        }
    \end{minipage}
    \vspace{-0.3cm}
    \caption{Effect of scaling the number of experts.}
    \label{tab:scaling_route_expert}
\end{subtable}%
\hfill
\begin{subtable}[t]{0.53\textwidth}
    \vspace{0pt}
    \centering
    \begin{minipage}[t][2.6cm][t]{\linewidth}
        \vspace{0pt}
        \centering
        \setlength{\tabcolsep}{3.5pt}
        \resizebox{\linewidth}{!}{%
        \begin{tabular}{lcccc}
            \toprule
            Method & \#PS & \#Params & \#Active & FID$\downarrow$ \\
            \midrule
            Dense-L/16 & 16 & 625M & 625M & 3.15 \\
            \rowcolor{rowblue}
            \textbf{PixelMoE-L-E32A4/16} & 16 & 1.65B & 634M & \textbf{2.89} \\
            \midrule
            Dense-L/8 & 8 & 622M & 622M & 2.95 \\
            \rowcolor{rowblue}
            \textbf{PixelMoE-L-E32A4/8} & 8 & 1.65B & 632M & \textbf{1.78} \\
            \bottomrule
        \end{tabular}%
        }
    \end{minipage}
    \vspace{-0.3cm}
    \caption{Computation scaling via token granularity.}
    \label{tab:moe_flops}
\end{subtable}
\caption{
    \textbf{Sparse scaling analysis of PixelMoE.}
    \#PS denotes the patch size.
}
\label{tab:sparse_scaling}
\end{table*}

\begin{figure*}[t]
    \centering
    \includegraphics[width=1.0\textwidth]{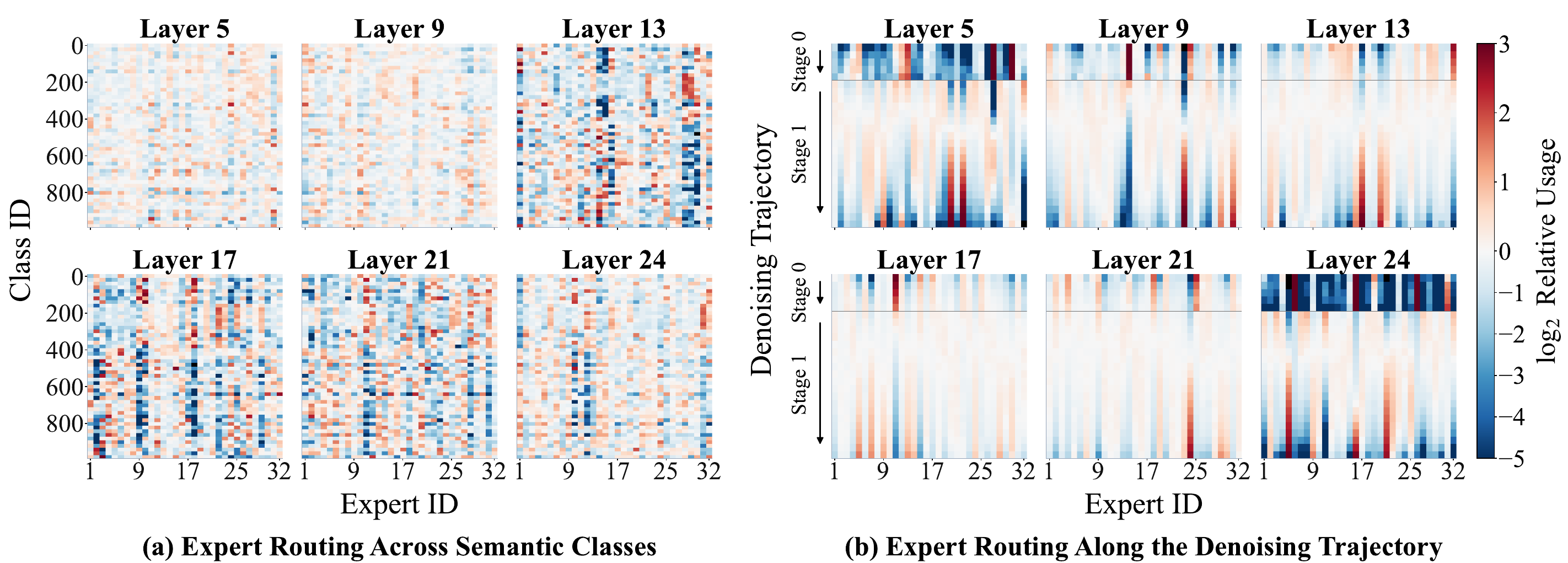}
    \caption{
    \textbf{Expert routing across semantic classes and along the denoising trajectory.}
    The left panel shows expert routing across ImageNet-1K classes, while the right panel shows how routing evolves along the denoising trajectory.
    }
    \label{fig:routing_specialization}
\end{figure*}

\paragraph{Active Parameter Scaling.}
We first compare PixelMoE with its dense counterparts across model scales.
For controlled comparisons, each dense baseline uses the same hierarchical flow setup and training recipe as its PixelMoE counterpart, with sparse MoE layers replaced by dense FFNs at comparable active parameter counts.
As shown in~\Cref{tab:activate_scaling}, PixelMoE consistently achieves lower FID than dense models across Base, Large, and Huge scales.
Beyond generation quality, we compare the convergence of Dense-L and PixelMoE-L-E32A4, extending the training to $200$ epochs for this analysis.
As shown in~\Cref{fig:epoch_fid_large}, PixelMoE-L-E32A4 reaches the FID achieved by Dense-L at $200$ epochs after only about $71$ epochs, reducing the required training epochs by a factor of approximately $2.8$.
These results show that sparse MoE scaling improves generation quality while also accelerating convergence at comparable active parameter counts.

\paragraph{Expert Scaling.}
We next isolate the effect of sparse model capacity by varying the number of routed experts $E$ while fixing the number of activated experts to $A=4$.
As shown in~\Cref{tab:scaling_route_expert}, increasing $E$ from $8$ to $16$ and then $32$ improves FID from $2.28$ to $2.03$ and $1.78$, respectively, while keeping the active parameter count fixed at $632$M.
This shows that PixelMoE effectively benefits from increased total model capacity without increasing the number of active parameters.

\paragraph{Computation Scaling via Token Granularity.}
We further study how PixelMoE benefits from increased computation by reducing the patch size, which increases the token sequence length while keeping the backbone scale fixed.
As shown in~\Cref{tab:moe_flops}, PixelMoE consistently outperforms the dense models at both patch sizes, with a larger improvement at finer token granularity.
A possible explanation is that smaller patches produce more spatially localized token representations and provide more token-level routing decisions, enabling the router to exploit finer-grained spatial variation.

\subsection{Expert Specialization}
We analyze the routing behavior of PixelMoE-L-E32A4 on the ImageNet-1K validation set.
For selected MoE layers, we collect token-level routing assignments and examine how expert preferences vary across semantic classes and along the denoising trajectory.
We report the $\log_2$ ratio of the observed assignment frequency to that under uniform routing, where positive and negative values indicate preferred and suppressed experts, respectively.

As shown in~\Cref{fig:routing_specialization}, PixelMoE exhibits distinct routing patterns across semantic classes and along the denoising trajectory.
Class-dependent routing is relatively weak in shallow layers, becomes clearer in intermediate layers, and is most pronounced around Layers 12--20, where different ImageNet-1K classes favor different subsets of experts.
In contrast, routing along the denoising trajectory is already prominent in shallow layers, weakens in intermediate layers, and emerges strongly again in the final layer.
The routing pattern also changes noticeably across the transition between the two hierarchical stages.
Overall, class-dependent routing specialization is more pronounced in intermediate layers, whereas specialization along the denoising trajectory is stronger in shallow and final layers.

\begin{wrapfigure}{r}{0.54\textwidth}
\vspace{-1.2em}
\centering
\includegraphics[width=\linewidth]{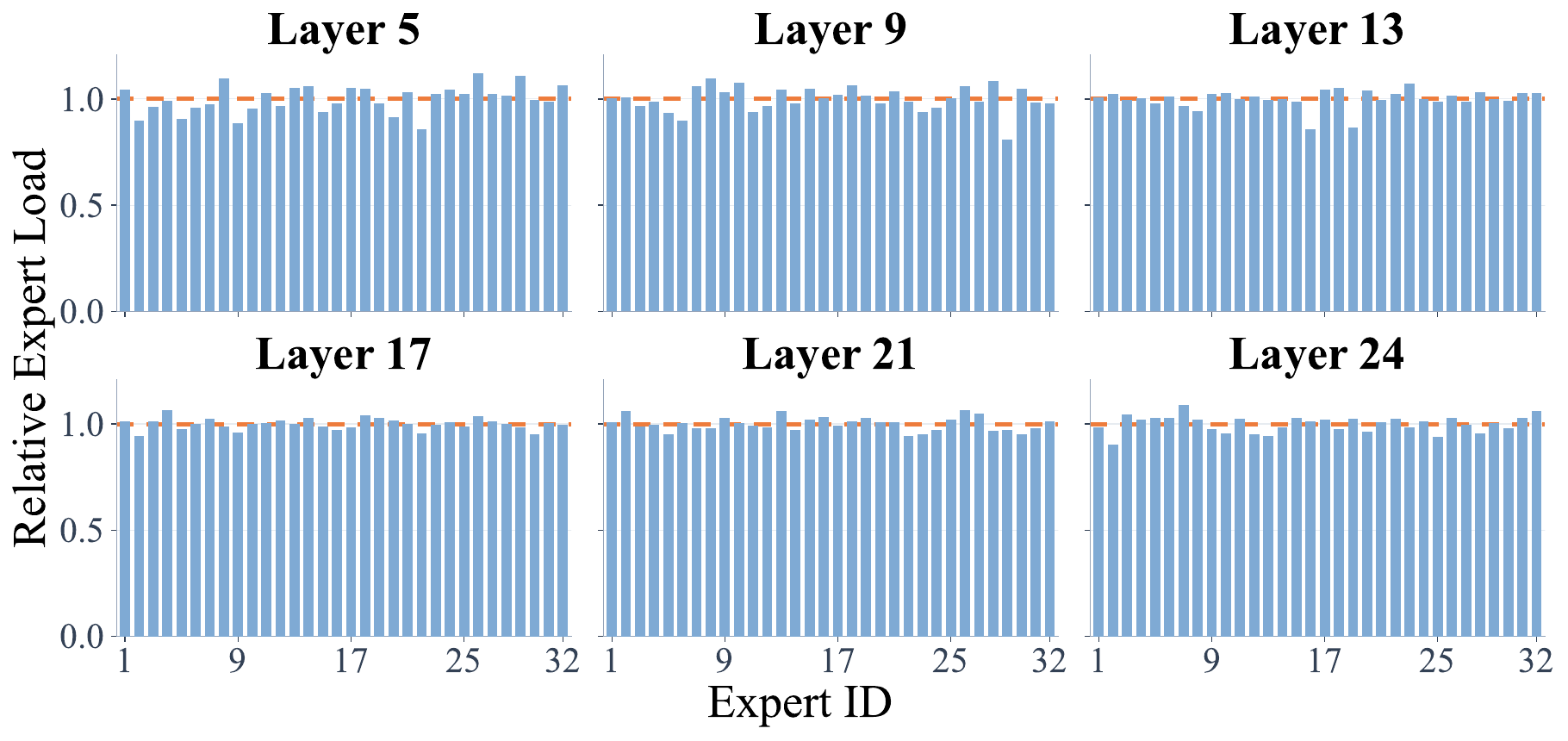}
\vspace{-0.5cm}
\caption{
\textbf{Global expert utilization across layers.}
The routed expert loads of PixelMoE-L-E32A4 remain close to a uniform distribution across the analyzed MoE layers.
Expert loads are normalized by their expected values under uniform routing, with the dashed line indicating 1.
}
\label{fig:routing_load}
\vspace{-1.0em}
\end{wrapfigure}
\paragraph{Balanced Global Utilization.}
The specialized routing patterns above could potentially arise from persistent over-utilization of a small subset of experts.
We therefore examine the global routed-expert loads across different MoE layers of PixelMoE-L-E32A4.
For each layer, we accumulate expert assignments over the ImageNet-1K validation set and report the resulting expert-load distribution.
As shown in~\Cref{fig:routing_load}, the routed
expert loads remain close to the uniform distribution across all analyzed layers.
This indicates that the bias-based load-balancing strategy effectively maintains globally balanced expert utilization while preserving input-adaptive expert specialization.
Together, these results show that the observed routing patterns across semantic classes and along the denoising trajectory do not simply reflect persistent over-utilization of a small subset of experts.

\subsection{Benchmarking against Previous Systems}
\begin{table*}[t]
\centering
\begin{adjustbox}{max width=\textwidth}
\setlength{\tabcolsep}{3pt}
\begin{tabular}{lccccccc}
\toprule
\multirow{2}{*}{Method} &
\multirow{2}{*}{Epochs} &
\multirow{2}{*}{\#Params} &
\multirow{2}{*}{\shortstack{\#Active\\Params}} &
\multicolumn{4}{c}{\textbf{ImageNet-1K Generation @ $256^2$}} \\
\cmidrule(lr){5-8}
&
&
&
&
FID$\downarrow$ &
IS$\uparrow$ &
Precision$\uparrow$ &
Recall$\uparrow$ \\
\arrayrulecolor{black}\midrule
\multicolumn{8}{l}{\textbf{\emph{Latent-space models: Dense}}}\\
\arrayrulecolor{black!30}\midrule
LDM-4-G~\citep{rombach2022high} & 170 & 400M+33M & 400M+33M & 3.60 & 247.6 & 0.87 & 0.48 \\
DiT-XL~\citep{dit} & 1400 & 675M+49M & 675M+49M & 2.27 & 278.2 & 0.83 & 0.57 \\
SiT-XL~\citep{ma2024sit} & 1400 & 675M+49M & 675M+49M & 2.06 & 270.3 & 0.82 & 0.59 \\
MaskDiT~\citep{zheng2024fast} & 1600 & 675M+49M & 675M+49M & 2.28 & 276.5 & 0.80 & 0.61 \\
REPA~\citep{yu2025representation} & 800 & 675M+49M & 675M+49M & 1.42 & 305.7 & 0.80 & 0.65 \\
SVG-XL~\citep{shi2026latent} & 1400 & 675M+43M & 675M+43M & 1.92  & 264.9 & -- & -- \\
FlowDCN-XL~\citep{wang2024exploring} & 400 & 618M+49M & 618M+49M & 2.00 & 263.2 & 0.82 & 0.58 \\
\arrayrulecolor{black}\midrule
\multicolumn{8}{l}{\textbf{\emph{Latent-space models: Sparse MoE}}}\\
\arrayrulecolor{black!30}\midrule
DiT-MoE-XL~\citep{fei2024scaling} & -- & 4.1B+49M  & 1.5B+49M  & 1.72 & 315.7 & 0.83 & 0.64 \\
RaceDiT-XL~\citep{yuan2025expert} &  1450 & 2.8B+49M  & 707M+49M  & 2.06 & 318.6& 0.83 &0.60 \\
DiffMoE~\citep{shi2025diffmoe} & 1400 & 1.2B+49M  & 458M+49M  & 2.13 & 274.4 & 0.81 & 0.60 \\
ProMoE-XL~\citep{wei2026routing} & 1000 & 1.6B+49M  & 675M+49M  & 2.59 & 265.6 & -- & -- \\
SharpMoE-L~\citep{deng2026focusing} & 1200 & 2.0B+49M  & 470M+49M  & 3.10  & 228.9 & -- & -- \\
\arrayrulecolor{black}\midrule
\multicolumn{8}{l}{\textbf{\emph{Pixel-space models: Dense}}}\\
\arrayrulecolor{black!30}\midrule
ADM~\citep{dhariwal2021diffusion} & 400 & 554M & 554M & 4.59  & 186.7 & 0.82 & 0.52 \\
RIN~\citep{rin} & 480 & 410M & 410M & 3.42 & 182.0 & -- & -- \\
Simple Diffusion~\citep{hoogeboom2023simple} & -- & 2.0B & 2.0B & 2.44 & 256.3 & -- & -- \\
JetFormer-L~\citep{tschannen2025jetformer} & 500 & 2.8B & 2.8B & 6.64 & -- & 0.69 & 0.56 \\
FractalMAR-H~\citep{li2025fractal} & 600 & 848M & 848M & 6.15 & 348.9 & 0.81 & 0.46 \\
PixelFlow-XL~\citep{chen2025pixelflow} & 320 & 677M & 677M & 1.98 & 282.1 & 0.81 & 0.60 \\
PixNerd-XL~\citep{wang2026pixnerd} & 320 & 700M & 700M & 1.93 & 298.0 & 0.80 & 0.60 \\
DiP-XL~\citep{chen2026dip} & 600 & 631M & 631M & 1.79 & 281.9 & 0.80 & 0.63 \\
JiT-G~\citep{li2026back} & 600 & 2.0B & 2.0B & 1.82 & 292.6 & 0.79 & 0.62 \\
PixelGen-XL~\citep{ma2026pixelgen} & 160 & 676M & 676M & 1.83 & 293.6 & 0.79 & 0.63 \\
PixelREPA-H~\citep{shin2026representation} & 600 & 953M & 953M & 1.81 & 317.2 & -- & -- \\
DeCo-XL~\citep{ma2026deco} & 600 & 682M & 682M & 1.69 & 304.0 & 0.79 & 0.63 \\
PixelDiT-XL~\citep{yu2026pixeldit} & 320 & 797M & 797M & 1.61  & 292.7 & 0.78 & 0.64 \\
HyperDiT-H~\citep{he2026hyperdit} & 600 & 952M & 952M & 1.56  & 306.5 & 0.80 & 0.64 \\
AsymFlow-H~\citep{chen2026asymmetric} & 600 & 953M & 953M & 1.57 & -- & -- & -- \\
\arrayrulecolor{black}\midrule
\multicolumn{8}{l}{\textbf{\emph{Pixel-space models: Sparse MoE}}}\\
\arrayrulecolor{black!30}\midrule
JiTMoE~\citep{liu2025efficient} & 200 & 1.3B & 465M & 3.10  & 311.3
 & -- & -- \\
\textbf{PixelMoE-B-E32A4} & 320 & 461M & 225M & 1.88 & 298.5 & 0.79 & 0.61\\
\textbf{PixelMoE-L-E32A4} & 320 & 1.65B & 632M & 1.54 & 315.9 & 0.81 & 0.61 \\
\textbf{PixelMoE-H-E32A4} & 320 & 3.68B & 1.47B & \textbf{1.47} & 302.2 & 0.79 & 0.64 \\
\arrayrulecolor{black}\bottomrule
\end{tabular}
\end{adjustbox}
\caption{\textbf{Class-conditional generation results on ImageNet-1K at $256\times256$.}
For latent-space methods, parameter counts include the generative model and the tokenizer decoder used at inference.}
\label{tab:imagenet256}
\end{table*}

\begin{table*}[t]
\centering
\begin{adjustbox}{max width=\textwidth}
\setlength{\tabcolsep}{3pt}
\begin{tabular}{lccccccc}
\toprule
\multirow{2}{*}{Method} &
\multirow{2}{*}{Epochs} &
\multirow{2}{*}{\#Params} &
\multirow{2}{*}{\shortstack{\#Active\\Params}} &
\multicolumn{4}{c}{\textbf{ImageNet-1K Generation @ $512^2$}} \\
\cmidrule(lr){5-8}
&
&
&
&
FID$\downarrow$ &
IS$\uparrow$ &
Precision$\uparrow$ &
Recall$\uparrow$ \\
\arrayrulecolor{black}\midrule
\multicolumn{8}{l}{\textbf{\emph{Latent-space models: Dense}}}\\
\arrayrulecolor{black!30}\midrule
DiT-XL~\citep{dit} & 1400 & 675M+49M & 675M+49M & 3.04 & 240.8 & 0.84 & 0.54 \\
SiT-XL~\citep{ma2024sit} & 1400 & 675M+49M & 675M+49M & 2.62 & 252.2 & 0.84 & 0.57 \\
MaskDiT~\citep{zheng2024fast} & 1600 & 675M+49M  & 675M+49M  & 2.50  & 256.2 & 0.83 & 0.56 \\
REPA~\citep{yu2025representation} & 800 & 675M+49M & 675M+49M & 2.08 & 274.6 & 0.83 & 0.58 \\
FlowDCN-XL~\citep{wang2024exploring} & 400 & 618M+49M & 618M+49M & 2.44 & 252.8 & 0.84 & 0.54 \\
\arrayrulecolor{black}\midrule
\multicolumn{8}{l}{\textbf{\emph{Latent-space models: Sparse MoE}}}\\
\arrayrulecolor{black!30}\midrule
DiT-MoE-XL~\citep{fei2024scaling} & -- & 4.1B+49M  & 1.5B+49M  & 2.30 & 298.4 & 0.85 & 0.57 \\
\arrayrulecolor{black}\midrule
\multicolumn{8}{l}{\textbf{\emph{Pixel-space models: Dense}}}\\
\arrayrulecolor{black!30}\midrule
ADM~\citep{dhariwal2021diffusion} & 400 & 554M & 554M & 3.85 & 221.7 & 0.84 & 0.53 \\
RIN~\citep{rin} & 480 & 410M & 410M & 3.95 & 216.0 & -- & -- \\
Simple Diffusion~\citep{hoogeboom2023simple} & -- &2.0B & 2.0B & 3.02 & 248.7 & -- & -- \\
PixNerd-XL~\citep{wang2026pixnerd} & 320 & 700M & 700M & 2.84 & 245.6 & 0.80 & 0.59 \\
DiP-XL~\citep{chen2026dip} & 600 & 631M & 631M & 2.31 & 291.7 & 0.84 & 0.58 \\
PixelDiT-XL~\citep{yu2026pixeldit} & 320 & 797M & 797M & 1.81 & 278.6 & 0.78 & 0.67 \\
JiT-G~\citep{li2026back} & 600 & 2.0B & 2.0B & 1.78 & 306.8 & -- & --\\
DeCo-XL~\citep{ma2026deco} & 340 & 682M & 682M & 2.22 & 290.0 & 0.80  & 0.60 \\
\arrayrulecolor{black}\midrule
\multicolumn{8}{l}{\textbf{\emph{Pixel-space models: Sparse MoE (Ours)}}}\\
\arrayrulecolor{black!30}\midrule
\textbf{PixelMoE-B-E32A4} & 320 & 463M & 227M  & 2.05 & 317.3 & 0.80 & 0.61 \\
\textbf{PixelMoE-L-E32A4} & 320 & 1.65B & 634M & 1.58 & 330.5 & 0.81 & 0.61 \\
\textbf{PixelMoE-H-E32A4} & 320 & 3.68B & 1.47B &  \textbf{1.53} & 322.2 & 0.80 & 0.63 \\
\arrayrulecolor{black}\bottomrule
\end{tabular}
\end{adjustbox}
\caption{\textbf{Class-conditional generation results on ImageNet-1K at $512\times512$.}
For latent-space methods, parameter counts include the generative model and the tokenizer decoder used at inference.}
\label{tab:imagenet512}
\end{table*}

\begin{figure*}[t]
    \centering
    \includegraphics[width=1.0\linewidth]{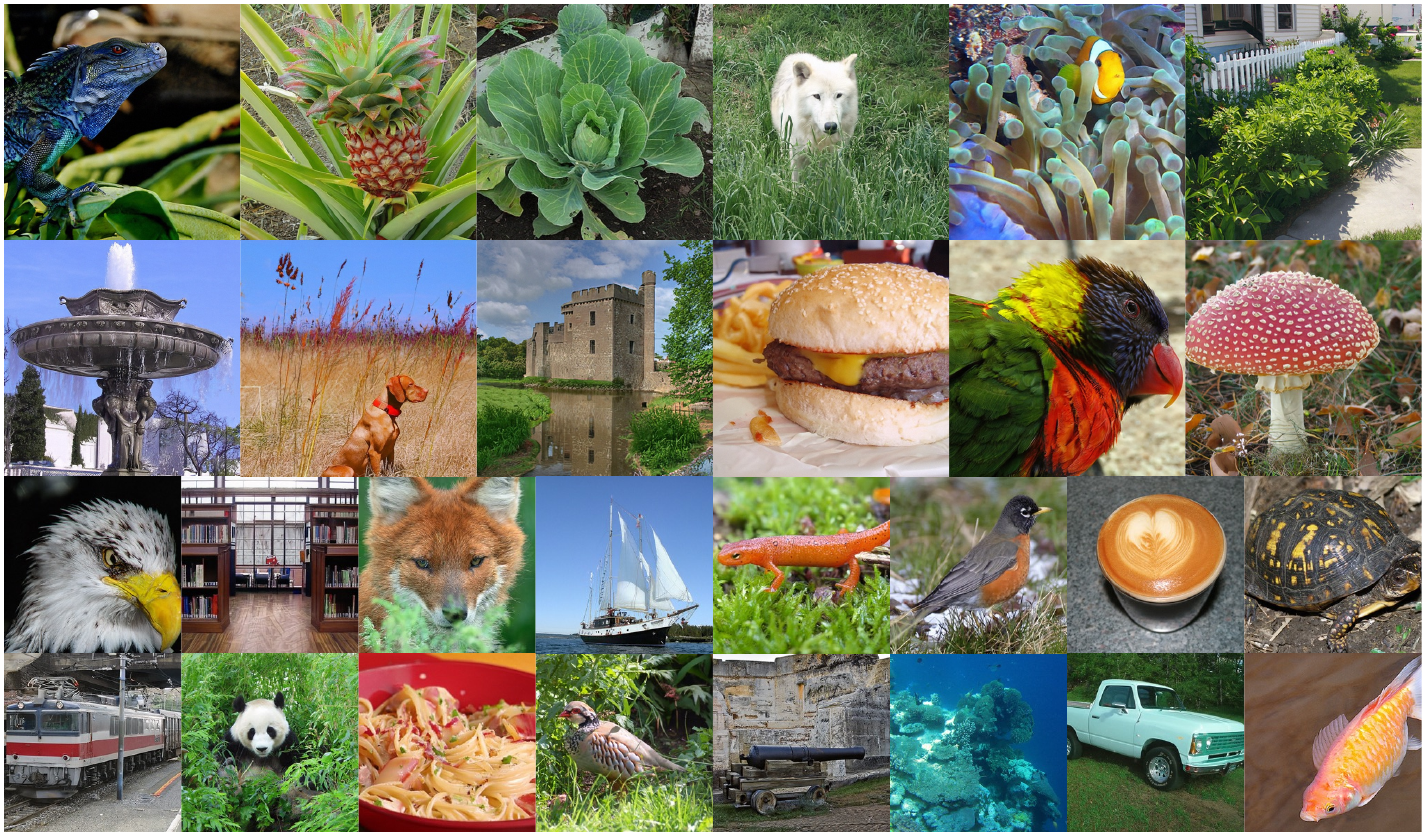}
    \caption{
    \textbf{Qualitative results of PixelMoE.} 
    Samples at $256\times256$ and $512\times512$ resolutions.
    }
    \label{fig:visualization}
\end{figure*}

\Cref{tab:imagenet256} compares PixelMoE with representative latent-space and pixel-space generative systems on class-conditional ImageNet-1K at $256\times256$ resolution.
PixelMoE-B-E32A4 achieves an FID of $1.88$ with only $225$M active parameters.
As total model capacity increases, PixelMoE-L-E32A4 further improves the FID to $1.54$, while PixelMoE-H-E32A4 reaches $1.47$.
Notably, these results are achieved without representation alignment~\citep{yu2025representation,shin2026representation} or perceptual losses~\citep{ma2026pixelgen}.
Compared with leading dense pixel-space models, PixelMoE achieves better generation quality.
PixelMoE also remains competitive with leading latent-space approaches without relying on a learned VAE.
Together with the controlled scaling results in~\Cref{tab:activate_scaling,tab:scaling_route_expert}, these comparisons support LLM-style sparse MoE as an effective scaling paradigm for pixel-space generative modeling.

We next examine whether this sparse scaling behavior of PixelMoE extends to higher-resolution pixel-space generation.
As shown in~\Cref{tab:imagenet512}, all PixelMoE models retain the same E32A4 sparse MoE design at $512\times512$ resolution.
PixelMoE-B-E32A4 achieves an FID of $2.05$, while PixelMoE-L-E32A4 and PixelMoE-H-E32A4 further improve the FID to $1.58$ and $1.53$, respectively.
Despite the substantially increased dimensionality of the pixel-space generation problem, PixelMoE maintains strong generation performance across model scales and remains competitive with representative dense pixel-space and latent-space generative models.
These results further demonstrate that the sparse scaling behavior of PixelMoE extends from $256\times256$ to higher-resolution pixel-space generation.

Qualitative results are shown in~\Cref{fig:visualization}.
PixelMoE generates coherent images across diverse ImageNet-1K categories, with clear global structures and rich local textures.
These samples further illustrate the strong visual fidelity of PixelMoE.

\subsection{Ablation Studies}
\label{sec:ablation}
We conduct ablation studies on the key components of PixelMoE.
The sparse MoE design and explicit routing guidance are evaluated on PixelMoE-L-E32A4, while the hierarchical flow schedule, stage-aware timestep sampling, and DDT prediction head are evaluated on PixelMoE-B-E32A4.
Unless otherwise specified, the experiments follow the default training recipe; for the hierarchical flow comparison, we adjust the number of training iterations to match the total number of processed image tokens.

\begin{table*}[t]
\centering
\small

\begin{subtable}[t]{0.45\textwidth}
    \vspace{0pt}
    \centering
    \begin{minipage}[t][2.0cm][t]{\linewidth}
        \vspace{0pt}
        \centering
        \resizebox{\linewidth}{!}{%
        \begin{tabular}{lccc}
            \toprule
            Exp. & \#Params & \#Active & FID$\downarrow$ \\
            \midrule
            w/o Shared Expert & 2.00B & 632M & 2.22 \\
            \rowcolor{rowblue}
            \textbf{w/ Shared Expert} & 1.65B & 632M & \textbf{1.78} \\
            \bottomrule
        \end{tabular}%
        }
    \end{minipage}
    \vspace{-0.6cm}
    \caption{Effect of the shared expert.}
    \label{tab:shared_expert_ablation}
\end{subtable}%
\hfill
\begin{subtable}[t]{0.54\textwidth}
    \vspace{0pt}
    \centering
    \begin{minipage}[t][2.0cm][t]{\linewidth}
        \vspace{0pt}
        \centering
        \resizebox{\linewidth}{!}{%
        \begin{tabular}{lcccc}
            \toprule
            Exp. & \#Expert Comb. & \#Params & \#Active & FID$\downarrow$ \\
            \midrule
            PixelMoE-L-E8A1 & 8 & 1.65B & 632M & 1.95 \\
            \rowcolor{rowblue}
            \textbf{PixelMoE-L-E32A4} & 35,960 & 1.65B & 632M & \textbf{1.78} \\
            \bottomrule
        \end{tabular}%
        }
    \end{minipage}
    \vspace{-0.6cm}
    \caption{Ablation on fine-grained routed experts.}
    \label{tab:activate_expert_ablation}
\end{subtable}

\par\medskip

\begin{subtable}[t]{0.45\textwidth}
    \vspace{0pt}
    \centering
    \begin{minipage}[t][2.2cm][t]{\linewidth}
        \vspace{0pt}
        \centering
        \resizebox{\linewidth}{!}{%
        \begin{tabular}{lccc}
            \toprule
            Exp. & \#Params & \#Active & FID$\downarrow$ \\
            \midrule
            Auxiliary Loss & 1.65B & 632M & 2.05 \\
            \rowcolor{rowblue}
            \textbf{Balancing Bias} & 1.65B & 632M & \textbf{1.78} \\
            \bottomrule
        \end{tabular}%
        }
    \end{minipage}
    \vspace{-0.6cm}
    \caption{Comparison of load-balancing strategies.}
    \label{tab:balancing}
\end{subtable}%
\hfill
\begin{subtable}[t]{0.54\textwidth}
    \vspace{0pt}
    \centering
    \begin{minipage}[t][2.2cm][t]{\linewidth}
        \vspace{0pt}
        \centering
        \resizebox{\linewidth}{!}{%
        \begin{tabular}{lcccc}
            \toprule
            Exp. & Type & \#Params & \#Active & FID$\downarrow$ \\
            \midrule
            Dense-L & Dense & 622M & 622M & 2.95 \\
            DiffMoE-L-E32A4 & EC & 1.67B & 653M & 2.62 \\
            \rowcolor{rowblue}
            \textbf{PixelMoE-L-E32A4} & TC & 1.65B & 632M & \textbf{1.78} \\
            \bottomrule
        \end{tabular}%
        }
    \end{minipage}
    \vspace{-0.6cm}
    \caption{Token-Choice \textit{vs.} Expert-Choice routing.}
    \label{tab:tc_ec}
\end{subtable}

\caption{
    \textbf{Ablation studies of the sparse MoE design in PixelMoE-L-E32A4.}
}
\label{tab:moe_ablation}
\end{table*}

\paragraph{The Shared Expert.}
We first examine the role of the shared expert in~\Cref{tab:shared_expert_ablation}.
Under matched active parameter counts, introducing a shared expert improves FID from $2.22$ to $1.78$, while reducing the total parameter count from $2.00$B to $1.65$B.
This suggests that combining shared and routed computation is more effective than allocating the entire active parameter budget to routed experts.

\paragraph{Fine-Grained Routed Experts.}
We next study the effect of expert granularity under matched total and active parameter counts.
As shown in~\Cref{tab:activate_expert_ablation}, changing the expert configuration from E8A1 to E32A4 expands the number of possible expert combinations from $8$ to $35{,}960$ without increasing either total or active parameter count.
The corresponding FID improvement from $1.95$ to $1.78$ indicates that PixelMoE benefits from finer-grained expert composition.

\paragraph{Bias-Based Load-Balancing Strategy.}
The bias-based load-balancing strategy offers a simple but effective way to regulate expert utilization without introducing an auxiliary loss.
As shown in~\Cref{tab:balancing}, the bias-based strategy improves FID from $2.05$ to $1.78$ compared with an auxiliary load-balancing loss, with identical total and active parameter counts.
Together with the expert-utilization analysis in~\Cref{fig:routing_load}, this result indicates that PixelMoE can preserve flexible input-adaptive routing while maintaining balanced global expert utilization.

\paragraph{Token-Choice Routing Strategy.}
We compare Token-Choice and Expert-Choice routing to examine which routing paradigm is better suited to pixel-space diffusion.
We implement DiffMoE~\citep{shi2025diffmoe}, which adopts Expert-Choice routing, using the same pixel-space backbone and training recipe as PixelMoE while keeping the total and active parameter counts comparable.
As shown in~\Cref{tab:tc_ec}, Token-Choice routing achieves a substantially lower FID.
This suggests that token-centric expert selection better accommodates the heterogeneous pixel-token representations across semantic classes and along the denoising trajectory.

\begin{table*}[t]
\centering
\small

\begin{subtable}[t]{0.47\textwidth}
    \vspace{0pt}
    \centering
    \begin{minipage}[t][2.2cm][t]{\linewidth}
        \vspace{0pt}
        \centering
        \setlength{\tabcolsep}{5pt}
        \resizebox{\linewidth}{!}{%
        \begin{tabular}{lccc}
            \toprule
            Routing Strategy & \#Params & \#Active & FID$\downarrow$ \\
            \midrule
            + Timestep Routing & 1.65B & 633M & 2.10 \\
            + Semantic Routing & 1.65B & 632M & 1.85 \\
            \rowcolor{rowblue}
            \textbf{PixelMoE-L-E32A4} & 1.65B & 632M & \textbf{1.78} \\
            \bottomrule
        \end{tabular}%
        }
    \end{minipage}
    \vspace{-0.6cm}
    \caption{Explicit routing guidance.}
    \label{tab:add_guidance}
\end{subtable}%
\hfill
\begin{subtable}[t]{0.47\textwidth}
    \vspace{0pt}
    \centering
    \begin{minipage}[t][2.2cm][t]{\linewidth}
        \vspace{0pt}
        \centering
        \setlength{\tabcolsep}{5pt}
        \resizebox{\linewidth}{!}{%
        \begin{tabular}{lccc}
            \toprule
            Generation Schedule & \#Params & \#Active & FID$\downarrow$ \\
            \midrule
            Single-stage flow & 461M & 225M & 2.54 \\
            \rowcolor{rowblue}
            \textbf{Hierarchical flow} & 461M & 225M & \textbf{2.46} \\
            \bottomrule
        \end{tabular}%
        }
    \end{minipage}
    \vspace{-0.6cm}
    \caption{Hierarchical flow schedule.}
    \label{tab:hierarchical_ablation}
\end{subtable}

\par\medskip

\begin{subtable}[t]{0.47\textwidth}
    \vspace{0pt}
    \centering
    \begin{minipage}[t][2.0cm][t]{\linewidth}
        \vspace{0pt}
        \centering
        \setlength{\tabcolsep}{3pt}
        \resizebox{\linewidth}{!}{%
        \begin{tabular}{lccc}
            \toprule
            Timestep Sampling & \#Params & \#Active & FID$\downarrow$ \\
            \midrule
            Global Distribution & 461M & 225M & 2.65 \\
            \rowcolor{rowblue}
            \textbf{Stage-Aware Sampling} & 461M & 225M & \textbf{2.46} \\
            \bottomrule
        \end{tabular}%
        }
    \end{minipage}
    \vspace{-0.6cm}
    \caption{Timestep sampling strategy.}
    \label{tab:timestep_dist_ablation}
\end{subtable}%
\hfill
\begin{subtable}[t]{0.47\textwidth}
    \vspace{0pt}
    \centering
    \begin{minipage}[t][2.0cm][t]{\linewidth}
        \vspace{0pt}
        \centering
        \setlength{\tabcolsep}{5pt}
        \resizebox{\linewidth}{!}{%
        \begin{tabular}{lccc}
            \toprule
            Prediction Head & \#Params & \#Active & FID$\downarrow$ \\
            \midrule
            w/o DDT Head & 369M & 133M & 3.76 \\
            \rowcolor{rowblue}
            \textbf{w/ DDT Head} & 461M & 225M & \textbf{2.46} \\
            \bottomrule
        \end{tabular}%
        }
    \end{minipage}
    \vspace{-0.6cm}
    \caption{DDT prediction head.}
    \label{tab:ddt_head}
\end{subtable}

\caption{
    \textbf{Ablation studies of routing and generation components in PixelMoE.}
    Explicit routing guidance is evaluated on PixelMoE-L-E32A4, while the remaining ablations are conducted on PixelMoE-B-E32A4.
}
\label{tab:additional_ablation}
\end{table*}

\paragraph{Explicit Routing Guidance.}
A natural question is whether expert routing requires explicit timestep or semantic information.
For timestep-aware routing, we provide the timestep embedding directly to the router together with the token hidden states; for semantics-aware routing, we follow the semantic routing design of ProMoE~\citep{wei2026routing}.
As shown in~\Cref{tab:add_guidance}, neither variant improves upon the default PixelMoE router, which makes routing decisions solely from token hidden states.
Together with the expert specialization observed across semantic classes and along the denoising trajectory in~\Cref{fig:routing_specialization}, these results suggest that token hidden states in pixel-space diffusion already contain rich timestep and semantic cues that support effective routing without explicit guidance.

\paragraph{Hierarchical Flow Schedule.}
The pixel-space hierarchical flow schedule progressively changes the spatial granularity along the denoising trajectory.
We compare it with a single-stage counterpart that performs the entire denoising process at full resolution.
For a fair comparison, we adjust the number of training iterations such that both variants process the same total number of image tokens.
As shown in~\Cref{tab:hierarchical_ablation}, hierarchical flow improves FID from $2.54$ to $2.46$.
This result shows the benefit of adapting spatial granularity along the denoising trajectory by using coarser representations for early denoising and finer representations for later refinement.

\paragraph{Stage-Aware Timestep Sampling.}
The two hierarchical stages denoise at different spatial resolutions, motivating stage-aware timestep sampling.
The baseline uses a single global Logit-Normal distribution $(\mu_{\mathrm g},\delta_{\mathrm g})=(0.8,0.8)$ to sample timesteps over the complete denoising trajectory.
In contrast, our default strategy uses the same global distribution for stage allocation and independently samples the stage-relative timestep $\sigma$ within each stage from a Logit-Normal distribution with $(\mu_{\mathrm l},\delta_{\mathrm l})=(0.6,0.8)$.
As shown in~\Cref{tab:timestep_dist_ablation}, this stage-aware sampling strategy improves FID from $2.65$ to $2.46$, allowing the timestep distribution within each stage to be controlled independently of how frequently that stage is sampled.

\paragraph{DDT Prediction Head.}
We further ablate the Decoupled Diffusion Transformer (DDT) prediction head~\citep{wang2026ddt} on PixelMoE-B-E32A4.
Without DDT, the backbone features are mapped directly to the high-dimensional pixel-space prediction, whereas the default model introduces a wider prediction pathway after the backbone.
As shown in~\Cref{tab:ddt_head}, the DDT head improves FID from $3.76$ to $2.46$.
This result supports the use of the DDT prediction head in our final configuration.

\section{Conclusion}
This paper presents \textbf{PixelMoE}, a pixel-space image generation framework that sparsely scales model capacity by directly adopting an LLM-style Mixture-of-Experts (MoE) architecture.
We further incorporate Pixel-Space Hierarchical Flow Matching to progressively increase spatial resolution along the denoising trajectory.
Experiments on ImageNet-1K show that PixelMoE consistently outperforms its dense counterparts across model scales and achieves competitive performance relative to leading latent- and pixel-space generative models.
Our routing analyses further reveal clear expert specialization across semantic classes and along the denoising trajectory under standard Token-Choice routing, without explicit semantic or timestep routing guidance.
These results support LLM-style sparse MoE as an effective scaling paradigm for pixel-space generative modeling.

\bibliography{iclr2027_conference}
\bibliographystyle{iclr2027_conference}

\newpage
\appendix

\section{Additional Details of Bias-Based Load Balancing}
\label{app:bias_balancing}

We provide the detailed computation of the expert-load distribution used in the bias-based load-balancing update.
For image token $i$, the assignment to expert $e$ is defined as
\begin{equation}
f^{s}_{t,i,e}=\begin{cases}\frac{1}{A}, & e\in\mathcal{S}^{s}_{t,i},\\
0, & \text{otherwise},
\end{cases}
\end{equation}
where $\mathcal{S}^{s}_{t,i}$ is the set of $A$ routed experts selected for image token $i$.
Averaging the assignments over all $N$ image tokens in the training batch gives the observed load fraction $F_e$ for expert $e$:
\begin{equation}
F_e=\frac{1}{N}\sum_{(s,t,i)\in\mathcal{T}}f^{s}_{t,i,e},
\end{equation}
where $\mathcal{T}$ denotes the image-token indices in the batch.
Since each image token contributes a total assignment mass of one, $\sum_{e=1}^{E}F_e=1$.

The uniform load target is $\mathbf{Q}=(1/E,\ldots,1/E)$.
For the bias update in~\Cref{sec:moe}, RMS normalization is defined as
\begin{equation}
\operatorname{RMS}(\mathbf{z})=\sqrt{\frac{1}{E}\sum_{e=1}^{E}z_e^2}.
\end{equation}
Accordingly, the balancing-bias vector is updated as
\begin{equation}
\mathbf{b}\leftarrow\mathbf{b}-\gamma\frac{\mathbf{F}-\mathbf{Q}}{\operatorname{RMS}(\mathbf{F}-\mathbf{Q})}.
\end{equation}
We use $\gamma=0.001$ throughout training.
A small epsilon of $10^{-6}$ is added to the denominator for numerical stability.
The balancing bias is updated without gradient propagation and affects expert selection only; the original routing scores are used to compute the routing weights.

\section{Scale Transition in Pixel-Space Hierarchical Flow Matching}
\label{app:hierarchical_transition}
During sampling, simply upsampling the low-resolution endpoint produced by Stage 0 does not yield a valid Stage 1 starting state, since nearest-neighbor upsampling replicates the noise associated with each low-resolution pixel and consequently introduces local correlations that are absent from the Stage 1 starting distribution used during training.
We therefore follow Pyramidal Flow Matching~\citep{jin2025pyramidal} and construct a distribution-preserving transition from the Stage 0 endpoint $\hat{x}^{e_0}_\theta$ to the Stage 1 starting state $\hat{x}^{s_1}_\theta$.
Specifically, the Stage 0 endpoint and the predefined Stage 1 starting state are given by
\begin{equation}
\begin{aligned}
x^{e_0}&=(1-t_{e_0})\mathcal{D}(x_0)+t_{e_0}\epsilon_0,\\
x^{s_1}&=(1-t_{s_1})\widetilde{x}_0+t_{s_1}\epsilon_1,\qquad
\widetilde{x}_0=\mathcal{U}(\mathcal{D}(x_0)),
\end{aligned}
\label{eq:transition_target_dist}
\end{equation}
where $\epsilon_0,\epsilon_1\sim\mathcal{N}(0,\delta_n^2 I)$.

Assuming an ideal Stage 0 prediction, upsampling its endpoint gives
\begin{equation}
\mathcal{U}(x^{e_0})=(1-t_{e_0})\widetilde{x}_0+t_{e_0}\mathcal{U}(\epsilon_0).
\end{equation}
To match the signal component of the predefined Stage 1 starting state, we first rescale the upsampled endpoint:
\begin{equation}
\frac{1-t_{s_1}}{1-t_{e_0}} \mathcal{U}(x^{e_0})=(1-t_{s_1})\widetilde{x}_0+\beta\,\mathcal{U}(\epsilon_0),
\end{equation}
where
\begin{equation}
\label{eq:def_beta}
\beta=\frac{1-t_{s_1}}{1-t_{e_0}}t_{e_0}.
\end{equation}
The signal coefficient now matches that of $x^{s_1}$, while the remaining difference lies in the noise covariance.

For $2\times$ nearest-neighbor upsampling, each low-resolution noise value is copied to a $2\times2$ block.
Therefore, after factoring out the noise variance $\delta_n^2$, the within-block covariance of $\mathcal{U}(\epsilon_0)$ is
\begin{equation}
\Sigma_{\mathrm{up}} =
\begin{bmatrix}
1 & 1 & 1 & 1\\
1 & 1 & 1 & 1\\
1 & 1 & 1 & 1\\
1 & 1 & 1 & 1
\end{bmatrix}.
\end{equation}
In contrast, the noise component $t_{s_1}\epsilon_1$ of the Stage 1 starting state has covariance $t_{s_1}^2\delta_n^2 I_4$, where $I_4\in\mathbb{R}^{4\times4}$ denotes the identity matrix over the four pixels within each $2\times2$ block.
We therefore introduce a block-correlated Gaussian perturbation $\epsilon'$, sampled independently of the Stage 0 noise, and construct
\begin{equation}
\hat{x}^{s_1}_\theta = \frac{1-t_{s_1}}{1-t_{e_0}} \mathcal{U}\!\left(\hat{x}^{e_0}_\theta\right) + \alpha\epsilon',
\label{eq:transition_general}
\end{equation}
where
\begin{equation}
\epsilon'\sim \mathcal{N}(0,\delta_n^2\Sigma_\rho),
\end{equation}
and
\begin{equation}
\Sigma_\rho=
\begin{bmatrix}
1 & \rho & \rho & \rho\\
\rho & 1 & \rho & \rho\\
\rho & \rho & 1 & \rho\\
\rho & \rho & \rho & 1
\end{bmatrix}.
\label{eq:transition_covariance}
\end{equation}
For $\Sigma_\rho$ to define a valid covariance matrix, it must be positive semidefinite. 
Its eigenvalues are $1-\rho$ with multiplicity three and $1+3\rho$ with multiplicity one, which requires $-\frac{1}{3}\leq\rho\leq1$.
Since the additional noise is used to cancel the positive within-block correlations introduced by nearest-neighbor upsampling, we further restrict $\rho$ to $-\frac{1}{3}\leq\rho\leq0$.

The actual within-block covariance of the resulting noise component in~\Cref{eq:transition_general} is
\begin{equation}
\delta_n^2\left(\beta^2\Sigma_{\mathrm{up}}+\alpha^2\Sigma_\rho\right).
\end{equation}
Matching this covariance to the target covariance $\delta_n^2 t_{s_1}^2 I_4$ of the predefined Stage 1 starting distribution and canceling the common factor $\delta_n^2$ yields
\begin{equation} 
\beta^2\Sigma_{\mathrm{up}} + \alpha^2\Sigma_\rho = t_{s_1}^2 I_4.
\label{eq:transition_cov_match} 
\end{equation} 
Matching the diagonal and off-diagonal entries gives
\begin{align}
    \beta^2 + \alpha^2 &= t_{s_1}^2, \\
    \beta^2 + \rho \alpha^2 &= 0.
\end{align}
Solving the two constraints yields
\begin{equation}
    \alpha^2=\frac{t_{s_1}^2}{1-\rho},\qquad \beta^2=-\frac{\rho}{1-\rho} t_{s_1}^2.
    \label{eq:transition_solution}
\end{equation}

Within the admissible range $\rho\in[-1/3,0]$, a more negative $\rho$ decreases $\alpha^2$, i.e., the variance coefficient of the additionally injected noise, while increasing the proportion of noise inherited from the upsampled Stage 0 endpoint. 
To make the transition minimally perturbative to the Stage 0 prediction, we minimize the variance of the newly injected noise subject to the covariance-matching constraint.
Following Pyramidal Flow Matching~\citep{jin2025pyramidal}, we therefore choose the most negative admissible value, $\rho=-1/3$. 
Substituting it into~\Cref{eq:transition_solution} gives
\begin{equation}
    \alpha=\frac{\sqrt{3}}{2} t_{s_1}, \qquad \beta=\frac{1}{2} t_{s_1}.
    \label{eq:transition_alpha}
\end{equation}
Combining the expression for $\beta$ with its definition in~\Cref{eq:def_beta} determines the relation between the two stage-boundary noise levels: 
\begin{equation} 
\frac{1-t_{s_1}}{1-t_{e_0}}t_{e_0} = \frac{t_{s_1}}{2}, 
\end{equation} 
which gives 
\begin{equation} 
t_{e_0} = \frac{t_{s_1}}{2-t_{s_1}}. 
\label{eq:transition_boundary} 
\end{equation} 
Thus, the Stage 0 endpoint and Stage 1 starting noise levels cannot be chosen independently; specifying either one determines the other under the covariance-matching transition.

The resulting cross-stage transition used during sampling is therefore
\begin{equation}
\hat{x}^{s_1}_{\theta}=\frac{1-t_{s_1}}{1-t_{e_0}}\mathcal{U}\!\left(\hat{x}^{e_0}_{\theta}\right) +\frac{\sqrt{3}}{2}t_{s_1}\epsilon',
\qquad
\epsilon'\sim\mathcal{N}\!\left(0,\delta_n^2\Sigma_{-1/3}\right).
\label{eq:transition_final}
\end{equation}
This transition preserves the coarse structure predicted by Stage 0 while matching the noise statistics of the predefined Stage 1 starting distribution and minimizing the amount of newly injected noise under the covariance-matching constraint.

\paragraph{Sampling the Correlated Noise.}
The required block-correlated Gaussian noise can be sampled directly. 
For each $2\times2$ spatial block and channel, we draw $\mathbf z\sim\mathcal N(0,I_4)$ and compute $\bar z=\frac{1}{4}\mathbf{1}^\top\mathbf z$, where $\mathbf{1}\in\mathbb{R}^4$ denotes the all-ones vector.
Then we set
\begin{equation}
    \epsilon'=\delta_n\frac{2}{\sqrt3}\left(\mathbf z-\bar z\mathbf{1}\right).
\end{equation}
This construction satisfies
\begin{equation}
    \operatorname{Cov}(\epsilon')=\delta_n^2\Sigma_{-1/3},
\end{equation}
providing a simple implementation of the correlated re-noising used in the cross-stage transition.

\section{Pseudo-code of PixelMoE}
\label{app:pseudocode}

\begin{algorithm}[tp]
\caption{Sparse MoE Layer in PixelMoE: PyTorch-like Pseudo-code}
\label{alg:pixelmoe}
\begin{lstlisting}
class PixelMoELayer(nn.Module):
    def __init__(
        self, dim, routed_dim, shared_dim,
        E=32, A=4, gamma=1e-3
    ):
        self.router = nn.Linear(dim, E, bias=False)
        self.experts = nn.ModuleList(
            [Expert(routed_dim) for _ in range(E)])
        self.shared_expert = Expert(shared_dim)

        self.router_bias = torch.zeros(E)
        self.A = A
        self.router_scale = math.sqrt(A)
        self.gamma = gamma

    def forward(self, x):
        scores = torch.sigmoid(self.router(x))

        # Balancing bias is used only for expert selection
        expert_ids = torch.topk(
            scores + self.router_bias,
            self.A, dim=-1
        ).indices

        # Normalize the original routing scores
        weights = gather(scores, expert_ids)
        weights = weights / weights.sum(
            dim=-1, keepdim=True)
        weights = self.router_scale * weights

        routed = sparse_expert_forward(
            x, self.experts, expert_ids, weights)

        shared = self.shared_expert(x)
        return routed + shared

    @torch.no_grad()
    def update_router_bias(self, load):
        target = load.mean()
        delta = load - target
        self.router_bias -= (
            self.gamma * delta / rms(delta)
        )
\end{lstlisting}
\end{algorithm}

\begin{algorithm}[tp]
\caption{Pixel-Space Hierarchical Flow Matching: PyTorch-like Pseudo-code}
\label{alg:hierarchical_flow}
\begin{lstlisting}
class HierarchicalFlow(nn.Module):
    def __init__(self, model):
        self.model = model  # shared PixelMoE

    def training_loss(self, x, c):
        stage = sample_stage()

        if stage == 0:
            x_low = downsample(x)
            noise = delta_n * torch.randn_like(x_low)

            x_start = noise
            x_end = (1 - t_e0) * x_low + t_e0 * noise
            t_start, t_end = 1.0, t_e0

        else:
            x_coarse = upsample(downsample(x))
            noise = delta_n * torch.randn_like(x)

            x_start = (1 - t_s1) * x_coarse + t_s1 * noise
            x_end = x
            t_start, t_end = t_s1, 0.0

        # Stage-relative timestep
        sigma = sample_timestep(stage)
        x_sigma = sigma * x_start + (1 - sigma) * x_end
        # Corresponding global timestep
        t = t_end + sigma * (t_start - t_end)

        # Stage-wise endpoint prediction
        x_end_pred = self.model(x_sigma, t, c)

        sigma_safe = sigma.clamp_min(sigma_min)
        loss = ((x_end_pred - x_end) / sigma_safe) ** 2
        return loss.mean()

    def velocity(self, x, t, c, stage):
        if stage == 0:
            t_start, t_end = 1.0, t_e0
        else:
            t_start, t_end = t_s1, 0.0
        # Convert global timestep to stage-relative timestep
        sigma = (t - t_end) / (t_start - t_end)

        # Convert endpoint prediction to velocity
        x_end_pred = self.model(x, t, c)
        sigma_safe = sigma.clamp_min(sigma_min)
        return (x - x_end_pred) / sigma_safe

    @torch.no_grad()
    def sample(self, c):
        # Stage 0: low-resolution generation
        x = delta_n * torch.randn(low_res_shape)
        x = heun_solver(self.velocity, x, c, stage=0)

        # Distribution-preserving stage transition
        x = stage_transition(x)

        # Stage 1: full-resolution refinement
        x = heun_solver(self.velocity, x, c, stage=1)

        return x
\end{lstlisting}
\end{algorithm}

We provide simplified PyTorch-like pseudo-code for the two key components of PixelMoE: the sparse MoE layer in~\Cref{alg:pixelmoe} and the two-stage pixel-space hierarchical flow matching schedule in~\Cref{alg:hierarchical_flow}.
In~\Cref{alg:hierarchical_flow}, the \texttt{heun\_solver} evaluates \texttt{velocity} at global timesteps $t$ but updates the state using stage-relative increments $\Delta\sigma=\Delta t/(t_{s_s}-t_{e_s})$.

\FloatBarrier
\section{Model Configuration}
\begin{table}[t]
\centering
\small
\setlength{\tabcolsep}{2pt}
\begin{tabular}{lcccccccc}
\toprule
Model & Depth & Hidden Dim. & Heads & DDT Depth & DDT Dim. & Routed Dim. & Shared Dim. & \#Dense \\
\midrule
\textbf{PixelMoE-B-E32A4} & 12 & 768  & 12 & 2 & 1536 & 384 & 632  & 4 \\
\textbf{PixelMoE-L-E32A4 }& 24 & 1024 & 16 & 2 & 2048 & 512 & 836  & 4 \\
\textbf{PixelMoE-H-E32A4} & 32 & 1280 & 16 & 4 & 2560 & 640 & 1048 & 4 \\
\bottomrule
\end{tabular}
\caption{
\textbf{PixelMoE model configurations.}
DDT Dim. represents the hidden state dimension in the DDT head.
Routed Dim. and Shared Dim. denote the intermediate dimensions of the routed-expert and shared-expert feed-forward networks (FFNs), respectively.
\#Dense denotes the number of dense blocks in the PixelMoE backbone.
}
\label{tab:model_config}
\end{table}
We implement PixelMoE at Base, Large, and Huge scales, with the detailed configurations summarized in~\Cref{tab:model_config}. 
For $256\times256$ generation, all models use a patch size of $8$; we use a patch size of $16$ at $512\times512$.
All models adopt the E32A4 MoE configuration unless otherwise specified, where each image token activates $4$ experts from a pool of $32$ routed experts.
As the backbone scales from Base to Huge, we increase its depth and hidden dimension together with the capacity of both routed and shared experts. 
A wide DDT head is appended to each backbone for stage-wise endpoint prediction, using $2$, $2$, and $4$ blocks for the Base, Large, and Huge models, respectively. 
Across all scales, the first four backbone blocks remain dense, while the remaining backbone blocks adopt sparse MoE layers; the DDT head retains dense FFNs.

\paragraph{Active Parameter Counting.}
In MoE blocks, image tokens are processed by the sparse MoE FFN, while class tokens use a separate dense FFN.
We report the average active parameter count per token, weighted according to the respective numbers of image and class tokens during training.

\section{Additional Training Details} 
\label{app:training_details} 

We summarize the detailed training configurations of PixelMoE in~\Cref{tab:training_config}. 
Unless otherwise specified, the same training recipe is used across model scales.
For $512\times512$ generation, we increase the patch size from $8$ to $16$ and the noise scale from $1.0$ to $2.0$, while keeping the remaining optimization, hierarchical flow, and MoE settings unchanged.

\paragraph{Controlled Dense Baselines.}
For controlled comparisons, each dense baseline follows the same backbone configuration, hierarchical flow schedule, DDT prediction head, optimizer, timestep sampling, and training recipe as the corresponding PixelMoE model.
The sparse MoE layers are replaced by dense FFNs with comparable active parameter counts, while all other architectural and training settings are kept unchanged.

\paragraph{Regularization.}
For the final PixelMoE-H-E32A4 models, we introduce additional regularization after epoch $200$.
Following LlamaGen~\citep{sun2024autoregressive}, we use a crop range of $1.1$, where each training image is first resized to $1.1\times$ the target resolution and then randomly cropped to the target size.
We additionally apply dropout with a rate of $0.2$ to the output of each self-attention module and feed-forward network.
These modifications increase the diversity of training views and provide stronger regularization during the later stage of training.

\begin{table*}[t]
\centering
\small
\setlength{\tabcolsep}{4pt}
\renewcommand{\arraystretch}{1.08}
\begin{tabular}{lcccccc}
\toprule
& \multicolumn{3}{c}{$256\times256$} 
& \multicolumn{3}{c}{$512\times512$} \\
\cmidrule(lr){2-4}
\cmidrule(lr){5-7}
Configuration 
& Base & Large & Huge 
& Base & Large & Huge \\
\midrule

\multicolumn{7}{l}{\textit{Input and training}} \\
\arrayrulecolor{black!30}\midrule

Training resolution
& $256^2$ & $256^2$ & $256^2$
& $512^2$ & $512^2$ & $512^2$ \\

Patch size
& $8$ & $8$ & $8$
& $16$ & $16$ & $16$ \\

Noise scale
& $1.0$ & $1.0$ & $1.0$
& $2.0$ & $2.0$ & $2.0$ \\

Crop range
& -- & -- & $1.1^\dagger$
& -- & -- & $1.1^\dagger$ \\

Dropout
& $0$ & $0$ & $0.2^\dagger$
& $0$ & $0$ & $0.2^\dagger$ \\

Label dropout
& \multicolumn{6}{c}{$0.1$} \\

Epochs
& \multicolumn{6}{c}{$100$ (default for scaling/ablation/routing), $320$ (benchmarking)} \\

Global batch size
& \multicolumn{6}{c}{$1024$} \\

Random horizontal flip
& \multicolumn{6}{c}{$0.5$} \\

\arrayrulecolor{black}\midrule
\multicolumn{7}{l}{\textit{Optimization}} \\
\arrayrulecolor{black!30}\midrule

Optimizer
& \multicolumn{6}{c}{gMuon + AdamW} \\

Learning rate
& \multicolumn{6}{c}{$2\times10^{-4}$} \\

Warm-up epochs
& \multicolumn{6}{c}{$5$} \\

LR schedule
& \multicolumn{6}{c}{Constant + cosine tail} \\

Cosine decay epochs
& \multicolumn{6}{c}{$40$} \\

Minimum learning rate
& \multicolumn{6}{c}{$2\times10^{-5}$} \\

Weight decay
& \multicolumn{6}{c}{$0$} \\

gMuon momentum
& \multicolumn{6}{c}{$0.95$} \\

AdamW $(\beta_1,\beta_2)$
& \multicolumn{6}{c}{$(0.9,0.95)$} \\

Precision
& \multicolumn{6}{c}{bfloat16} \\

EMA decay
& \multicolumn{6}{c}{$0.9999,\ 0.9996$} \\

\arrayrulecolor{black}\midrule
\multicolumn{7}{l}{\textit{Hierarchical flow training}} \\
\arrayrulecolor{black!30}\midrule

Number of stages
& \multicolumn{6}{c}{$2$} \\

Stage resolutions
& $128^2, 256^2$ & $128^2, 256^2$ & $128^2, 256^2$
& $256^2, 512^2$ & $256^2, 512^2$ & $256^2, 512^2$ \\

Stage 0 endpoint $t_{e_0}$
& \multicolumn{6}{c}{$0.75$} \\

Stage 1 start $t_{s_1}$
& \multicolumn{6}{c}{$6/7$} \\

Noise correlation $\rho$
& \multicolumn{6}{c}{$-1/3$} \\

Global timestep distribution
& \multicolumn{6}{c}{Logit-Normal $(\mu_{\mathrm g}=0.8,\delta_{\mathrm g}=0.8)$} \\

Stage-relative distribution
& \multicolumn{6}{c}{Logit-Normal $(\mu_{\mathrm l}=0.6,\delta_{\mathrm l}=0.8)$} \\

Endpoint-loss clamp $\sigma_{\min}$
& \multicolumn{6}{c}{$0.05$} \\

\arrayrulecolor{black}\midrule
\multicolumn{7}{l}{\textit{MoE training}} \\
\arrayrulecolor{black!30}\midrule

Router score
& \multicolumn{6}{c}{Sigmoid} \\

Router precision
& \multicolumn{6}{c}{FP32} \\

Routing scale
& \multicolumn{6}{c}{$\sqrt{A}$} \\

Load balancing
& \multicolumn{6}{c}{Bias-based} \\

Bias update rate
& \multicolumn{6}{c}{$0.001$} \\

Bias update rule
& \multicolumn{6}{c}{RMS-normalized} \\

\bottomrule
\end{tabular}

\caption{
\textbf{Training configurations of PixelMoE.}
For $512\times512$ generation, we increase the patch size from $8$ to $16$ and the Gaussian noise scale from $1.0$ to $2.0$.
$^\dagger$ For the final PixelMoE-H-E32A4 models, LlamaGen-style crop augmentation with a crop range of $1.1$ and dropout with a rate of $0.2$ are applied after epoch $200$; the $100$-epoch controlled experiments use the standard preprocessing without these additional regularization measures.
}
\label{tab:training_config}
\end{table*}

\paragraph{Optimization.} 
We use gMuon~\citep{GramNewtonSchulz} for eligible matrix-valued parameters, including the routed-expert matrices, while AdaLN layers, the final prediction layer, timestep and class embedders, and the remaining parameters are optimized with AdamW~\citep{loshchilov2018decoupled}. 
We use a learning rate of $2\times10^{-4}$ with a global batch size of $1024$.
After a $5$-epoch warm-up, we keep the learning rate constant and apply cosine decay to $2\times10^{-5}$ over the final $40$ epochs. 
We train with $\mathtt{bfloat16}$ mixed precision and maintain two EMA copies with decay rates $0.9996$ and $0.9999$. 

\paragraph{Hierarchical Flow Training.} 
We use the same two-stage pixel-space hierarchical flow matching schedule for both resolutions, with Stage 0 denoising at half spatial resolution and Stage 1 at full resolution. 
We set the Stage 0 endpoint to $t_{e_0}=0.75$ and use $\rho=-1/3$ for the noise correlation. 
Under the cross-stage transition derived in~\Cref{app:hierarchical_transition}, these settings determine the Stage 1 starting noise level as $t_{s_1}=6/7$.
The relative sampling frequency of the two stages is determined by a global Logit-Normal distribution with $(\mu_{\mathrm{g}},\delta_{\mathrm{g}})=(0.8,0.8)$, with the probability mass in the overlapping stage interval divided equally between the two stages.
The stage-relative timestep is then independently sampled from a Logit-Normal distribution with $(\mu_{\mathrm{l}},\delta_{\mathrm{l}})=(0.6,0.8)$.
Under the resulting stage intervals, this yields sampling probabilities
$p(s=0)\approx0.231$ and $p(s=1)\approx0.769$.

\paragraph{Stage-wise Endpoint Prediction Objective.}
PixelMoE directly predicts the endpoint of each stage while using the velocity-space flow matching loss introduced in~\Cref{sec:hierarchical}.
For numerical stability near the stage endpoint, we clamp the inverse-square weighting using $\sigma_{\min}=0.05$.
The implemented objective is
\begin{equation}
\mathcal{L}_{\mathrm{v\text{-}endpoint}} =\mathbb{E}\left[\frac{\mathcal{L}_{\mathrm{MSE}}\left(\hat{x}^{e_s}_{\theta},x^{e_s}\right)}{\max(\sigma,\sigma_{\min})^2}\right],
\end{equation}
where the expectation is taken over images, sampled stages, stage-relative timesteps, and Gaussian noise.
The inverse-square weighting is not an additional loss-reweighting heuristic.
Before applying the numerical clamp, the endpoint parameterization gives
\begin{equation}
v_\theta-v^s=\frac{x^{e_s}-\hat{x}^{e_s}_\theta}{\sigma},
\end{equation}
so velocity-space MSE directly yields the $1/\sigma^2$ weighting on endpoint-prediction error.
The clamp at $\sigma_{\min}$ only prevents this factor from becoming singular near the stage endpoint.

\paragraph{MoE Training.} 
The router operates in FP32 and converts the routing logits into sigmoid scores. 
For each image token, we select the top-$A$ experts according to the bias-adjusted scores, normalize the original sigmoid scores over the selected experts, and multiply the resulting routing weights by a scale factor of $\sqrt{A}$ before expert aggregation. 
The balancing bias only affects expert selection and is not used when computing the expert routing weights. 
After each optimizer update, expert loads accumulated during the corresponding forward passes are used to update the balancing bias with an RMS-normalized correction and a step size of $0.001$. 
No auxiliary load-balancing objective is used for the default PixelMoE models.

\paragraph{Auxiliary Load-Balancing Baseline.}
For the auxiliary load-balancing loss baseline in~\Cref{tab:balancing}, we replace the bias-based balancing strategy with the Token-Choice load-balancing loss~\citep{fedus2022switch,liu2024deepseek}.
Specifically, let $r_{i,e}$ denote the sigmoid routing score of image token $i$ for expert $e$, and let $\mathcal{S}_i$ denote its selected top-$A$ experts.
We first normalize the routing scores over all experts as
\begin{equation}
    p_{i,e} =\frac{r_{i,e}}{\sum_{j=1}^{E} r_{i,j}}.
\end{equation}
For each expert, we define the mean routing probability and assignment fraction as
\begin{equation}
    P_e=\frac{1}{N}\sum_{i=1}^{N}p_{i,e},\qquad
    F_e=\frac{1}{NA}\sum_{i=1}^{N}\mathbf{1}[e\in\mathcal{S}_i],
\end{equation}
where $N$ is the number of image tokens.
The load-balancing loss for the $\ell$-th MoE layer is
\begin{equation}
\mathcal{L}_{\mathrm{bal}}^{(\ell)}=E\sum_{e=1}^{E}\operatorname{sg}(F_e)P_e,
\end{equation}
where $\operatorname{sg}(\cdot)$ denotes stop-gradient.
We average this loss across all MoE layers and add it to the training objective:
\begin{equation}
\mathcal{L}_{\mathrm{aux}}=\lambda_{\mathrm{aux}}\cdot\frac{1}{L_{\mathrm{MoE}}}\sum_{\ell=1}^{L_{\mathrm{MoE}}}\mathcal{L}_{\mathrm{bal}}^{(\ell)},
\end{equation}
where $L_{\mathrm{MoE}}$ denotes the number of MoE layers.
Following standard practice in prior MoE models~\citep{fedus2022switch,dai2024deepseekmoe}, we set $\lambda_{\mathrm{aux}}=0.01$.
For this baseline, the training objective is
\begin{equation}
\mathcal{L}=\mathcal{L}_{\mathrm{v\text{-}endpoint}}+\mathcal{L}_{\mathrm{aux}}.
\end{equation}

\section{Evaluation Details}
\label{app:evaluation_details}
We follow the evaluation setup of JiT~\citep{li2026back} for FID and IS, including its ImageNet-1K reference statistics, and use the ADM~\citep{dhariwal2021diffusion} implementation for Precision and Recall.
We summarize our sampling protocol in~\Cref{tab:evaluation_config}.
During sampling, we construct a 50-point uniform global timestep grid, include the stage boundaries, and use the grid points within the timestep interval of each stage.
For class-conditional ImageNet-1K generation, class labels are uniformly distributed over the 1,000 classes across distributed workers.

\begin{table}[t]
\centering
\small
\setlength{\tabcolsep}{7pt}
\renewcommand{\arraystretch}{1.08}
\begin{tabular}{lc}
\toprule
Configuration & Setting \\
\midrule
Sampling solver & Heun \\
Global timestep grid & $50$ \\
Global grid spacing & Uniform \\
Timestep shifting & None \\
\midrule
CFG scale & Swept per model \\
CFG interval & Swept per model \\
EMA candidates & $0.9996{,}\ 0.9999$ \\
Selection samples & $10{,}000$ \\
\midrule
Final evaluation samples & $50{,}000$ \\
Class sampling & Uniform over $1{,}000$ classes \\
Metrics & FID, IS, Precision, Recall \\
\bottomrule
\end{tabular}
\caption{
\textbf{Evaluation configurations.}
CFG scale, CFG interval, and EMA decay are jointly selected for each model according to FID-10K, and the selected configuration is fixed for the final FID-50K evaluation.
}
\label{tab:evaluation_config}
\end{table}

\paragraph{Classifier-Free Guidance.}
We apply classifier-free guidance~\citep{ho2021classifierfree} as
\begin{equation}
v_{\mathrm{cfg}}(x_\sigma^s,t,c)=v_\theta(x_\sigma^s,t,\phi)+w(t)\left(v_\theta(x_\sigma^s,t,c)-v_\theta(x_\sigma^s,t,\phi)\right),
\end{equation}
where $c$ and $\phi$ denote the class and null conditions, respectively, and $w(t)$ is controlled by the CFG scale $w$ and the CFG interval~\citep{kynkaanniemi2024applying}.
For an interval $[t_{\min},t_{\max}]$, we define
\begin{equation}
    w(t)=
    \begin{cases}
        w, & t_{\min} < t < t_{\max},\\
        1, & \text{otherwise}.
    \end{cases}
\end{equation}
Thus, CFG is applied only within the selected portion of the sampling trajectory.
Outside this interval, sampling uses the standard conditional prediction without CFG amplification.

\paragraph{Model Selection.}
For every evaluated model, we jointly sweep the CFG scale, CFG interval, and the two EMA checkpoints.
Each candidate configuration is first evaluated with 10,000 samples, and the configuration achieving the lowest FID-10K score is selected.
We then regenerate 50,000 samples using the selected configuration and report all final generation metrics on this sample set.

\section{More Visualization Results}
\label{app:more_visualization}

We provide additional \textit{uncurated} samples of PixelMoE-H-E32A4 on ImageNet-1K.
\Cref{fig:more_vis_256_1,fig:more_vis_256_2} show samples at $256\times256$, while \Cref{fig:more_vis_512} presents samples at $512\times512$.
All samples are generated conditioned on the specified classes using the same sampling settings as the reported quantitative results.

\newcommand{\addclasspdf}[3]{
\begin{minipage}[t]{0.49\linewidth}
\centering
\includegraphics[width=\linewidth]{#1}\\[-0.2em]
{\scriptsize {Class #2}: #3}
\end{minipage}
}

\begin{figure*}[t]
\centering

\addclasspdf{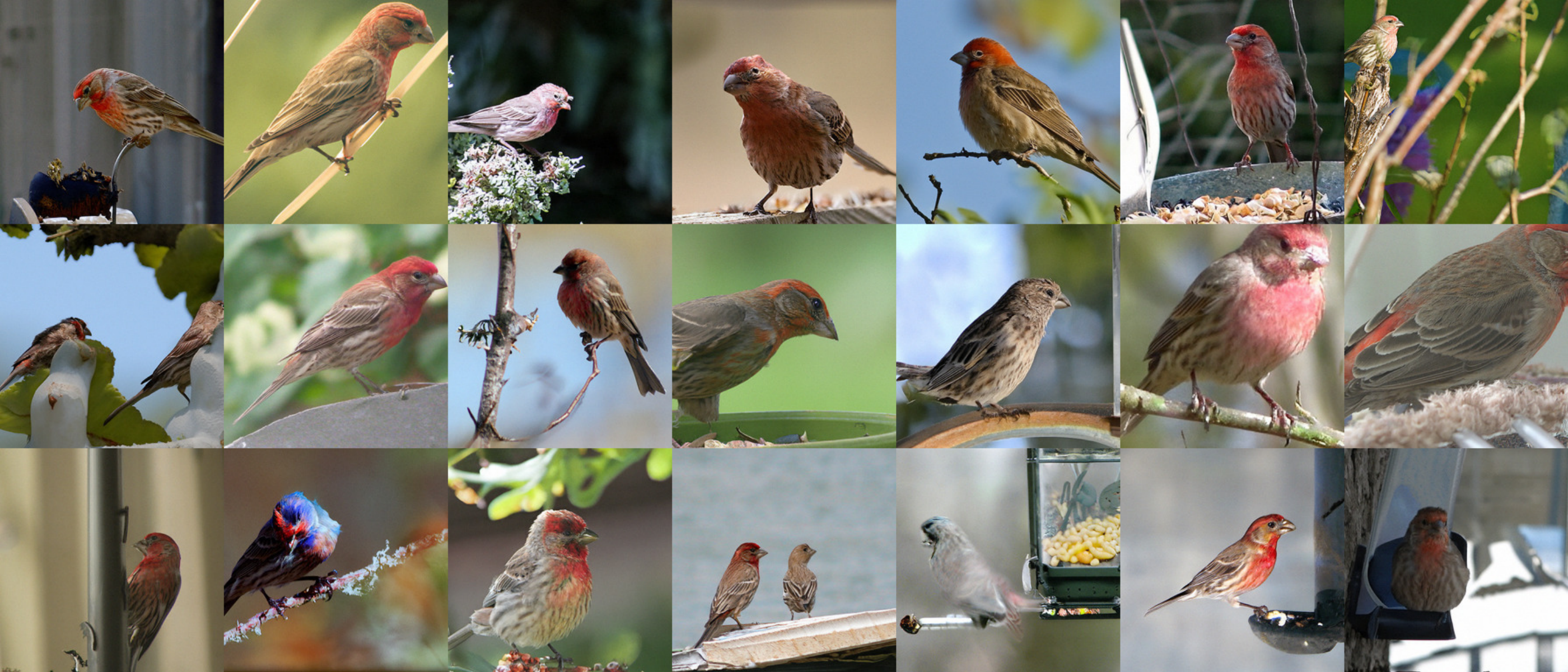}
{012}{House finch, Linnet, Carpodacus mexicanus}
\addclasspdf{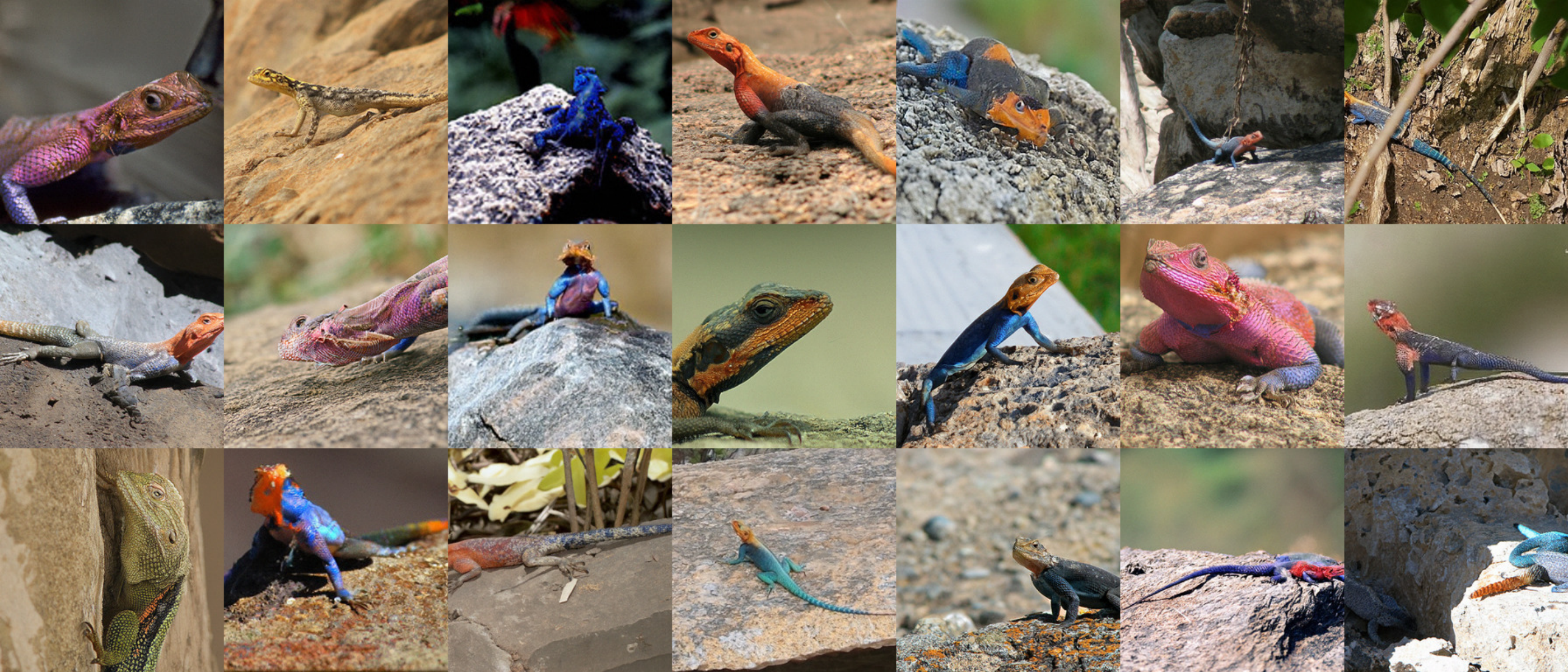}
{042}{Agama}
\\

\addclasspdf{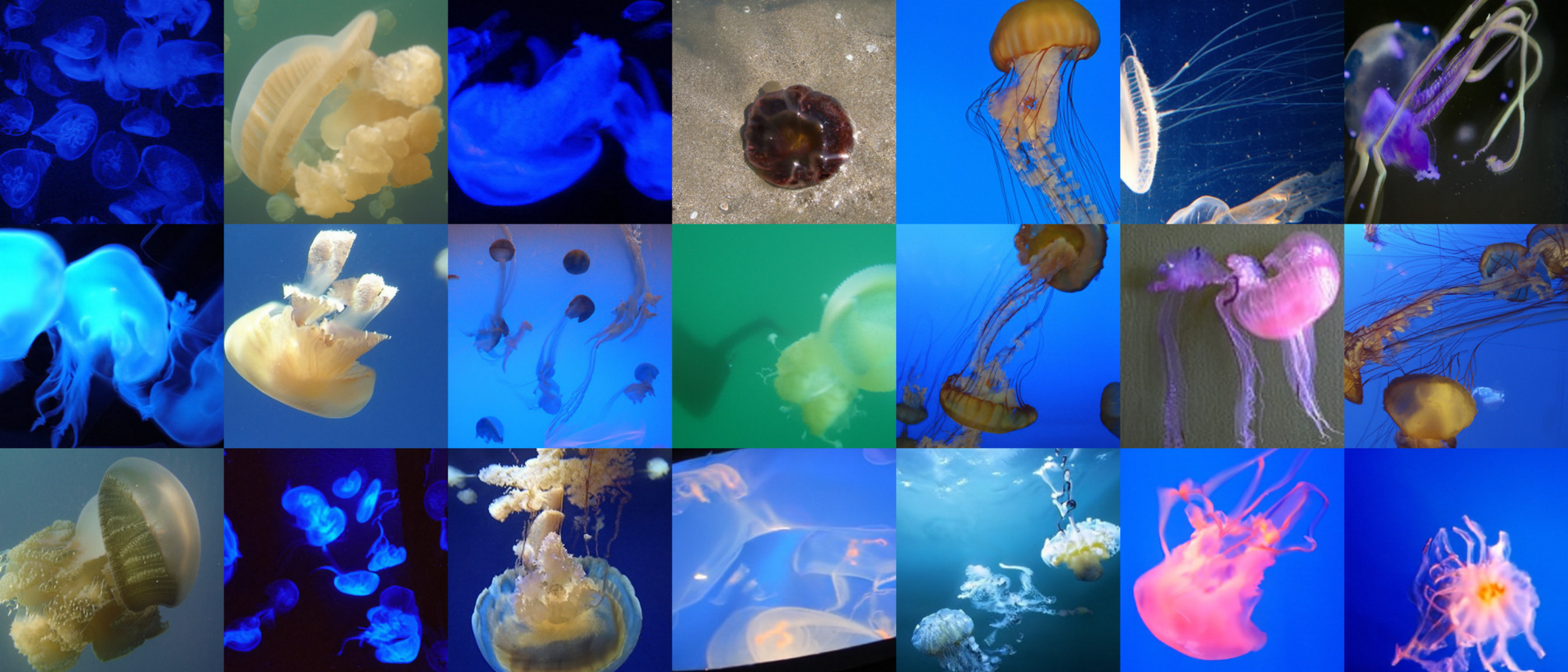}
{107}{Jellyfish}
\addclasspdf{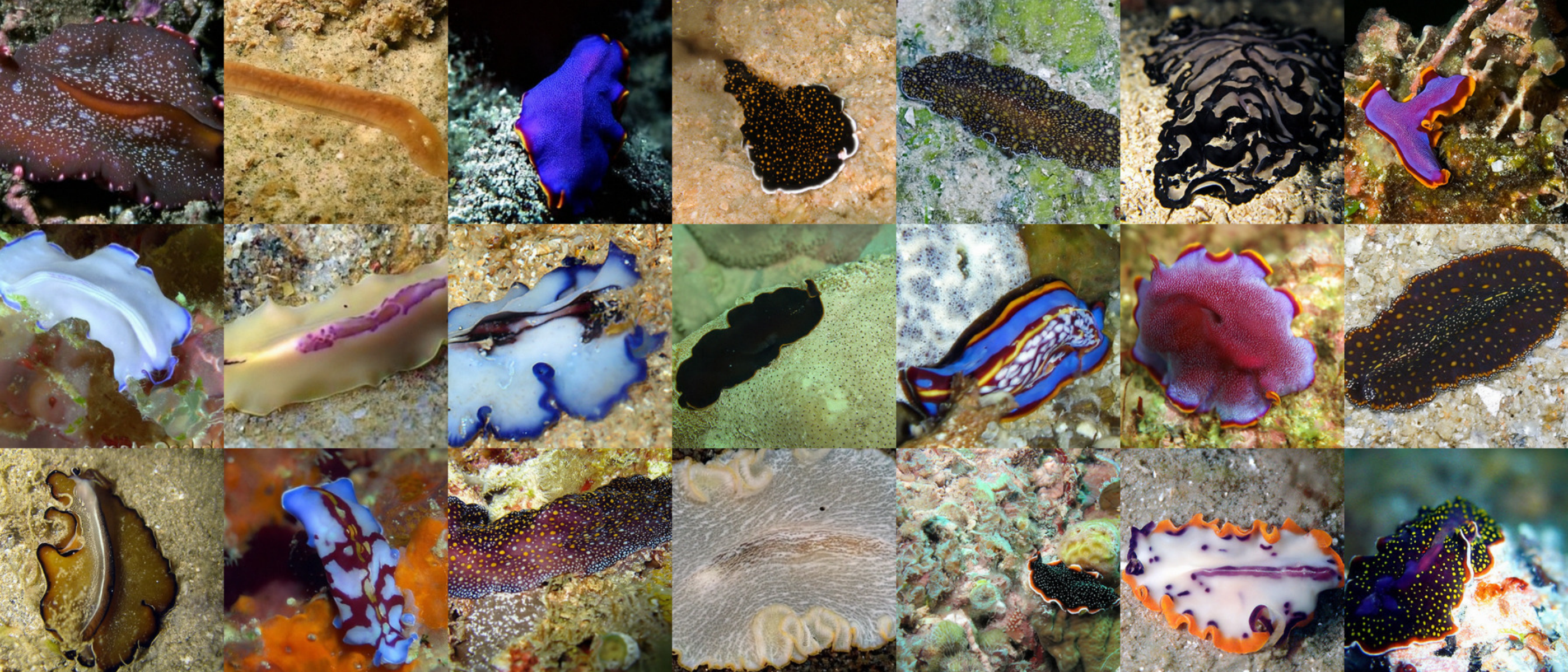}
{110}{Flatworm, Platyhelminth}
\\

\addclasspdf{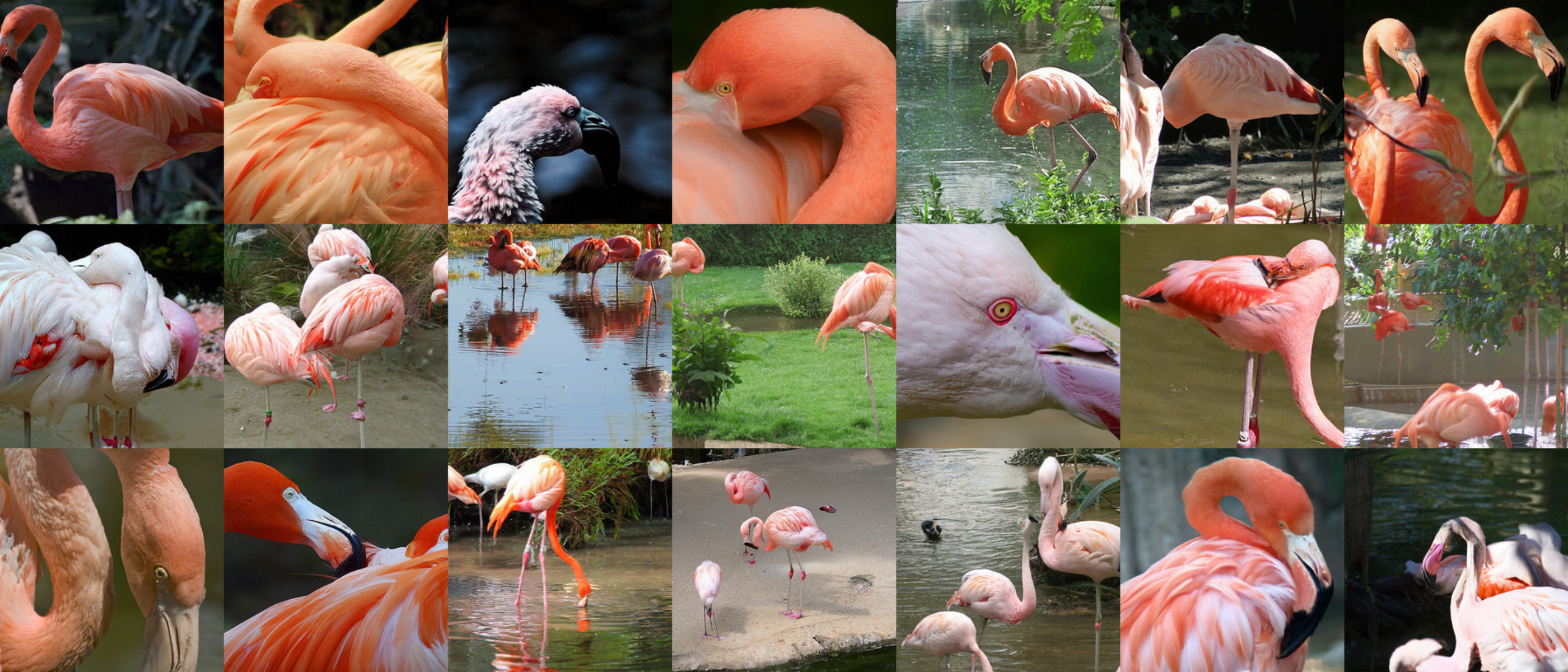}
{130}{Flamingo}
\addclasspdf{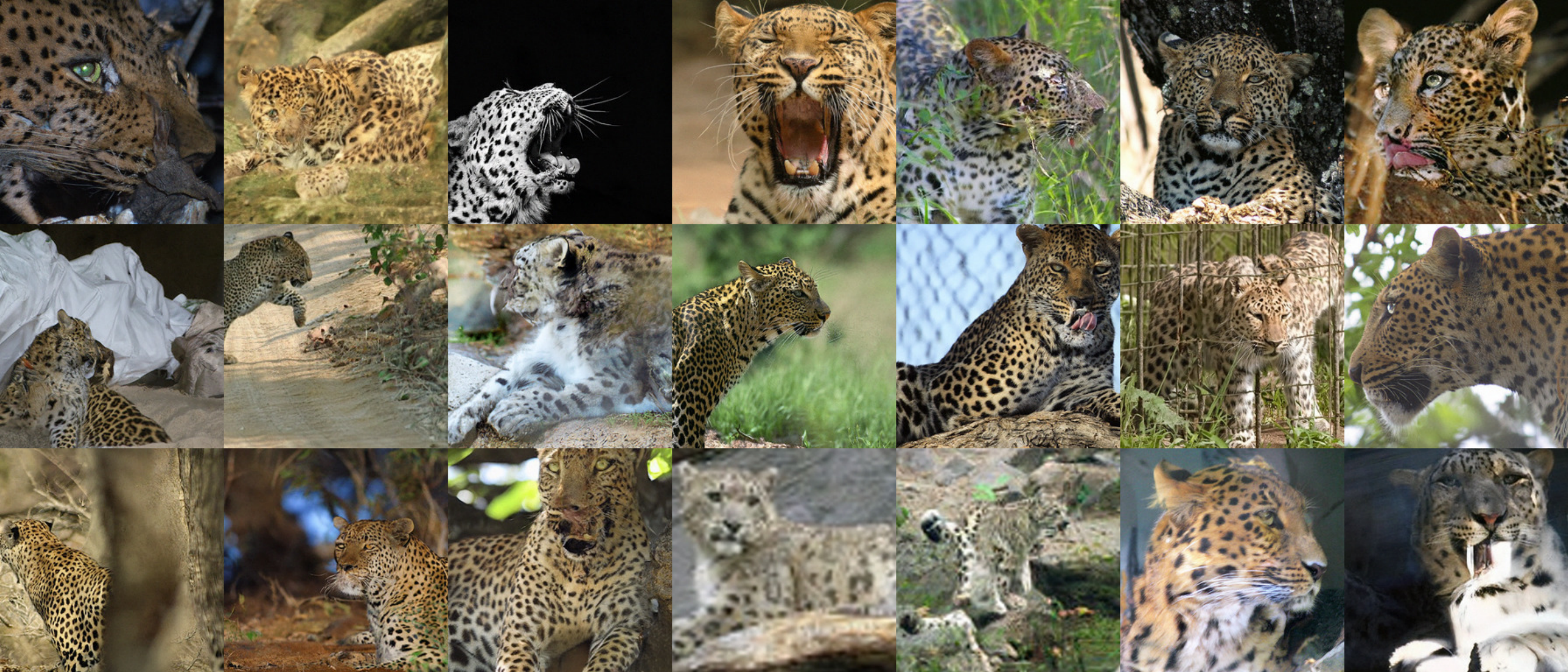}
{288}{Leopard, Panthera pardus}
\\

\addclasspdf{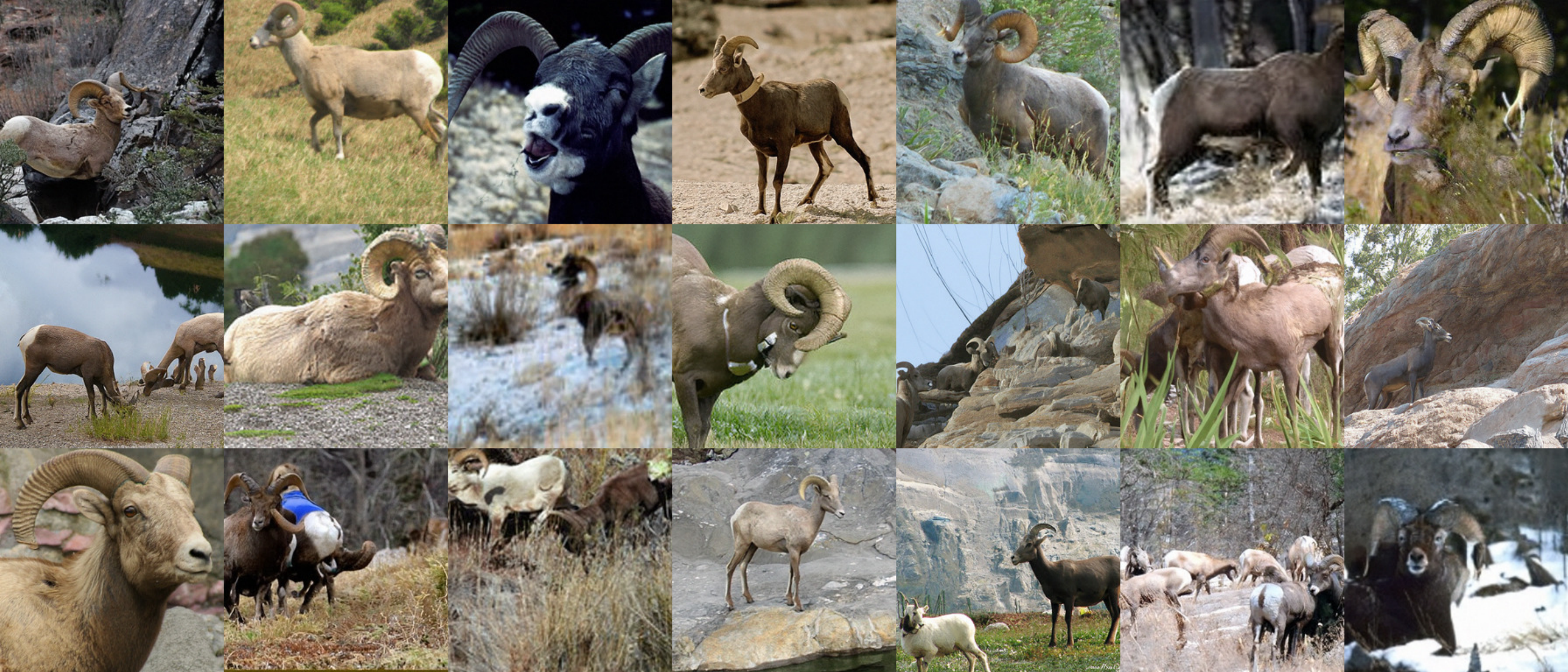}
{349}{Bighorn, Bighorn sheep, Cimarron, Rocky Mountain bighorn}
\addclasspdf{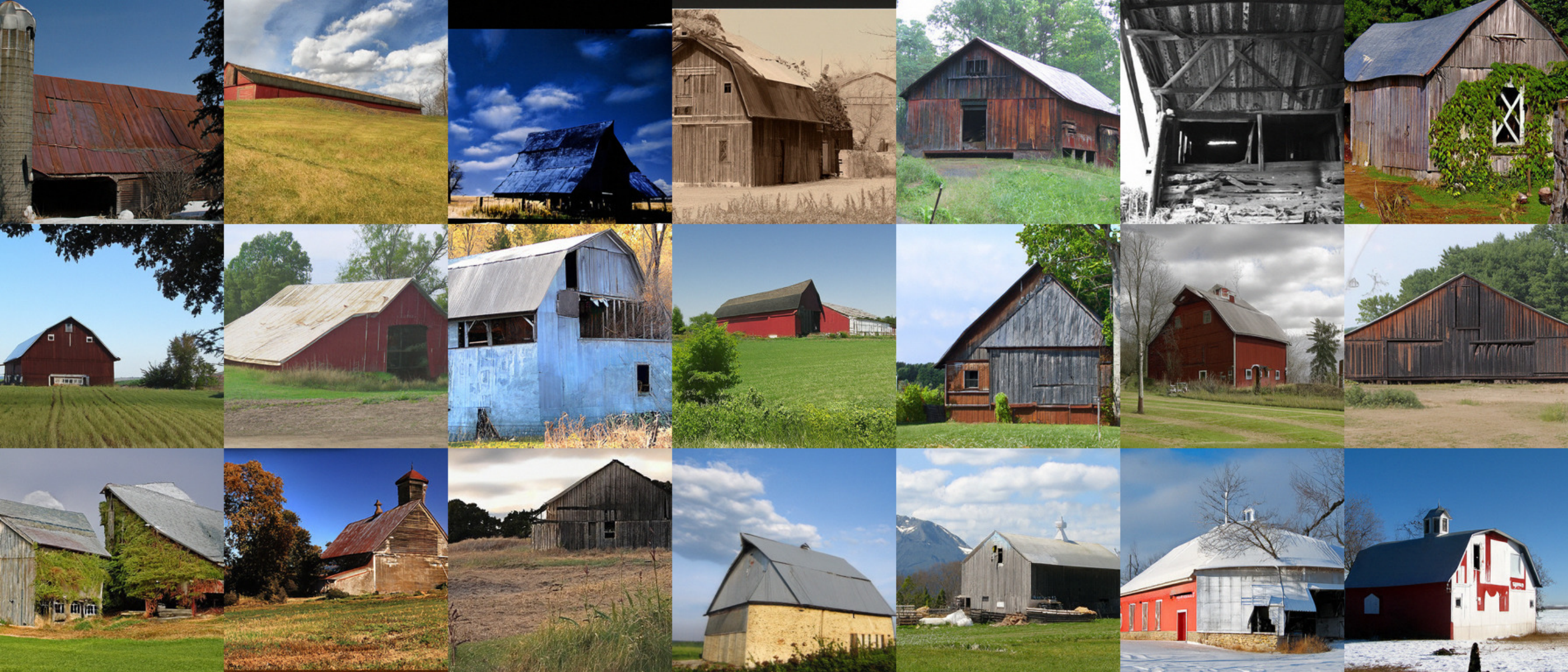}
{425}{Barn}
\\
\addclasspdf{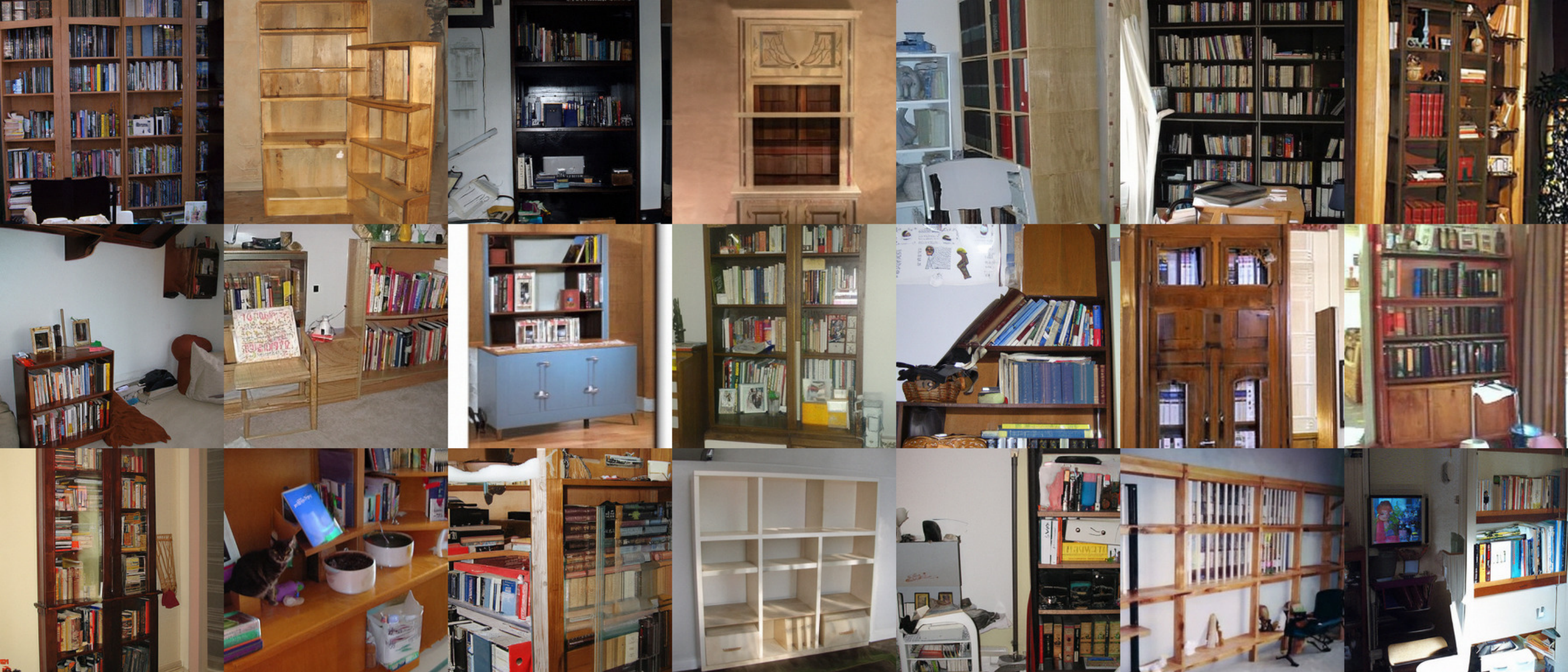}
{453}{Bookcase}
\addclasspdf{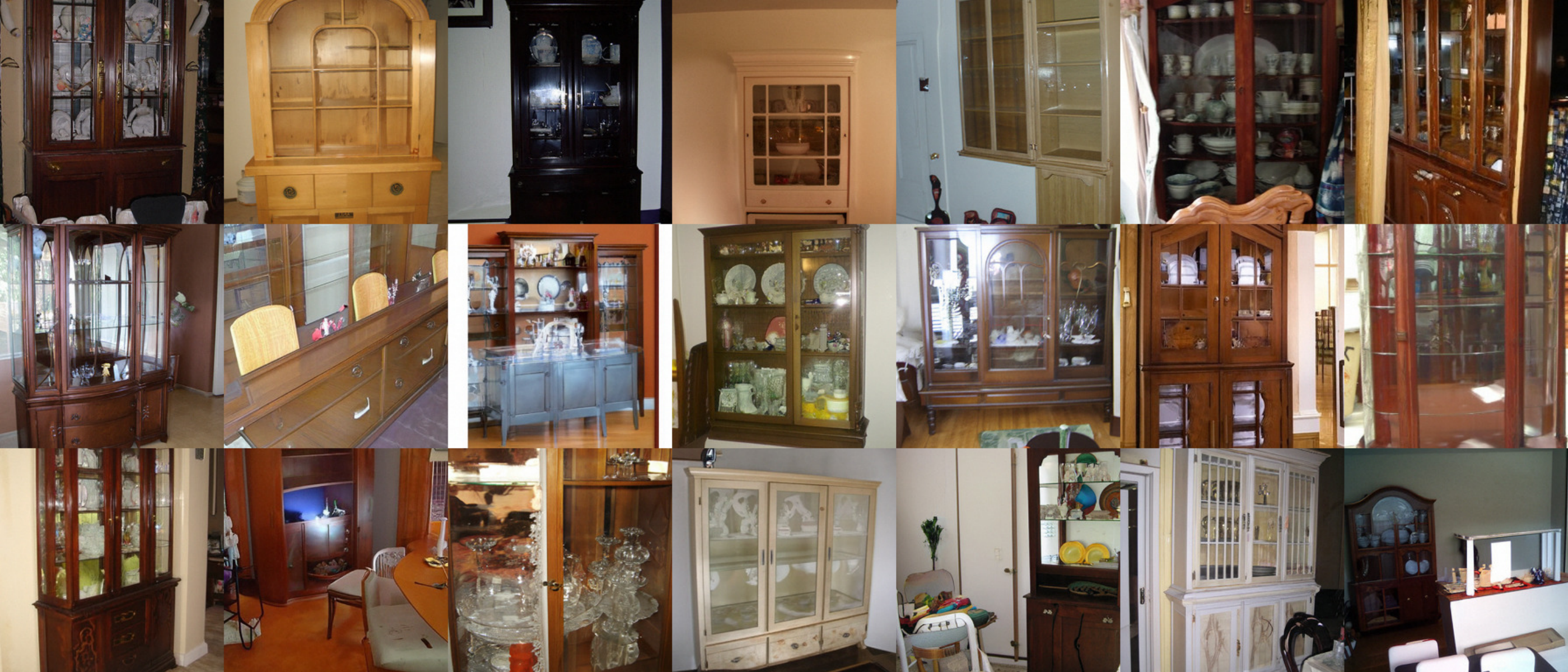}
{495}{China cabinet, China closet}
\\
\caption{
\textbf{\textit{Uncurated} samples of PixelMoE-H-E32A4 on ImageNet-1K at $256\times256$.}
Samples are generated conditioned on the specified classes using the same sampling configuration as the reported quantitative results.
}
\label{fig:more_vis_256_1}
\end{figure*}

\begin{figure*}[t]
\centering

\addclasspdf{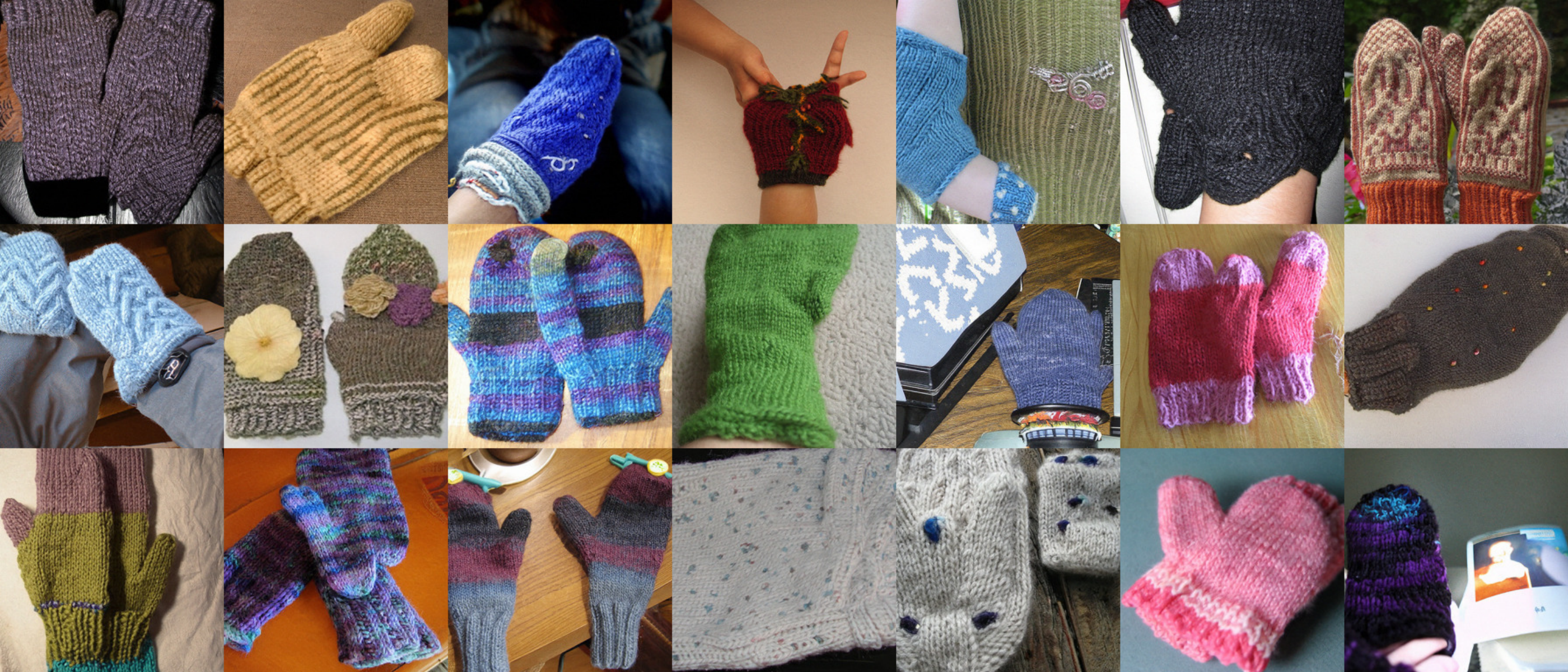}
{658}{Mitten}
\addclasspdf{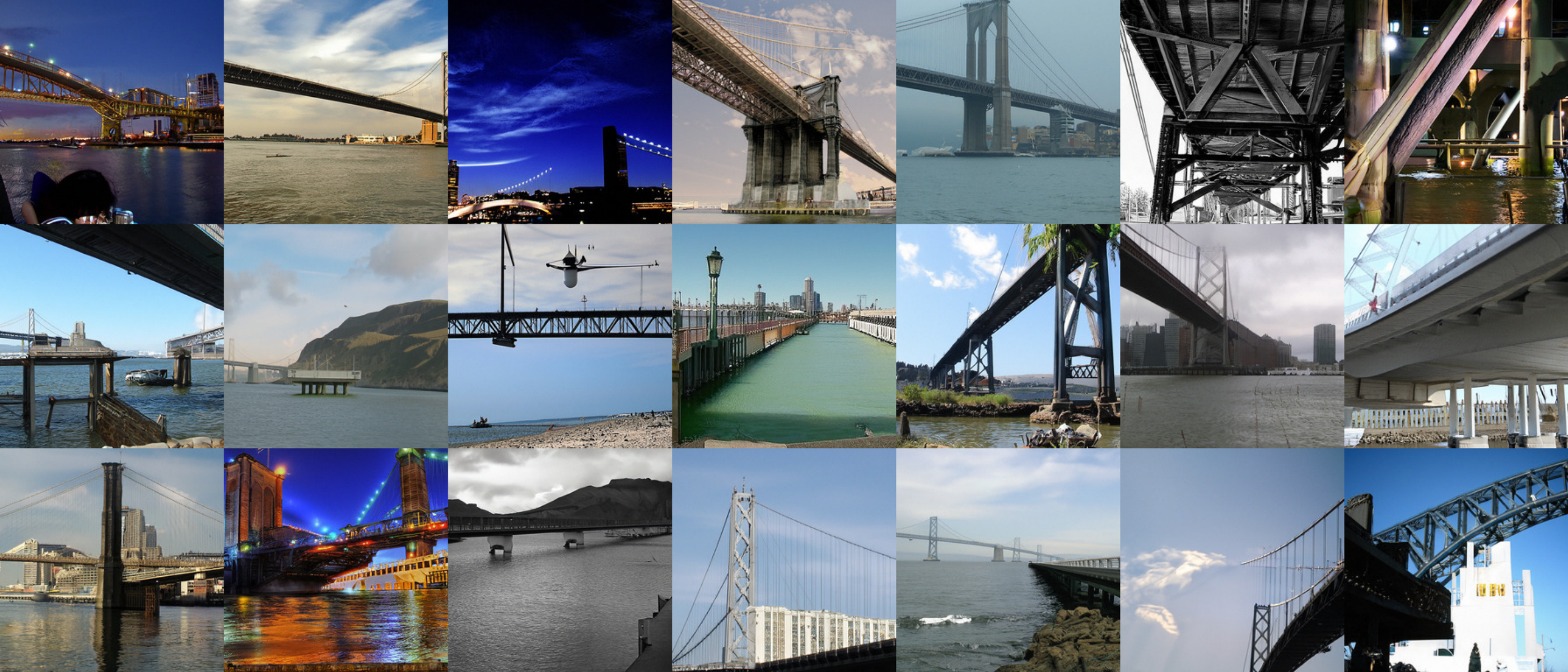}
{718}{Pier}
\\

\addclasspdf{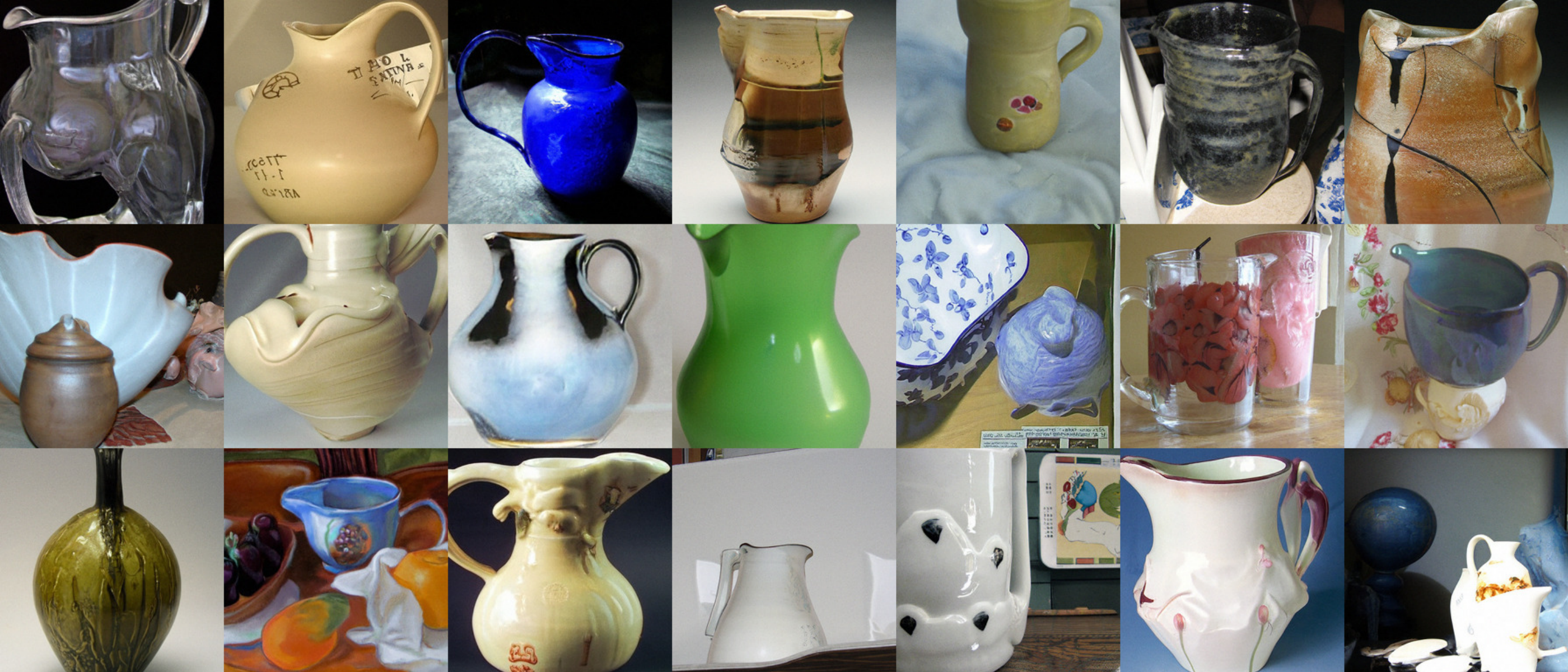}
{725}{Pitcher, Ewer}
\addclasspdf{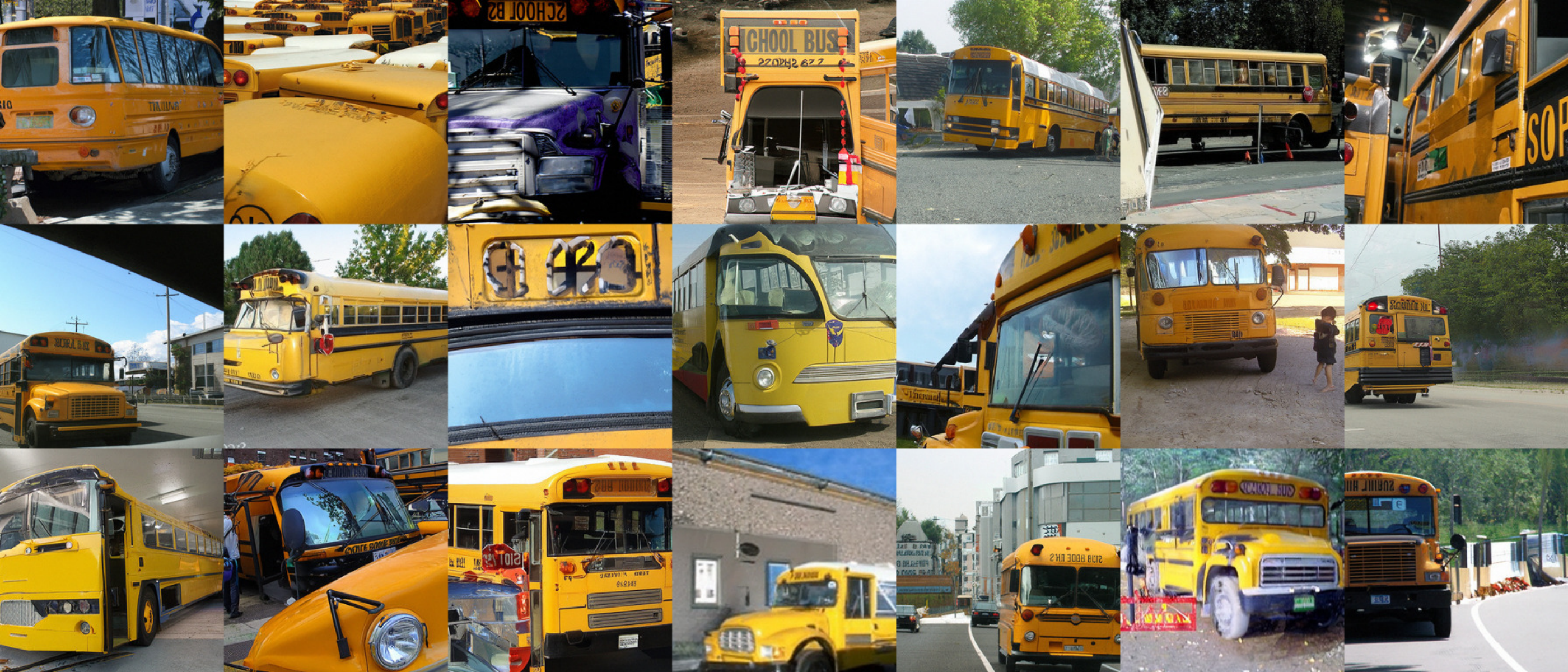}
{779}{School bus}
\\

\addclasspdf{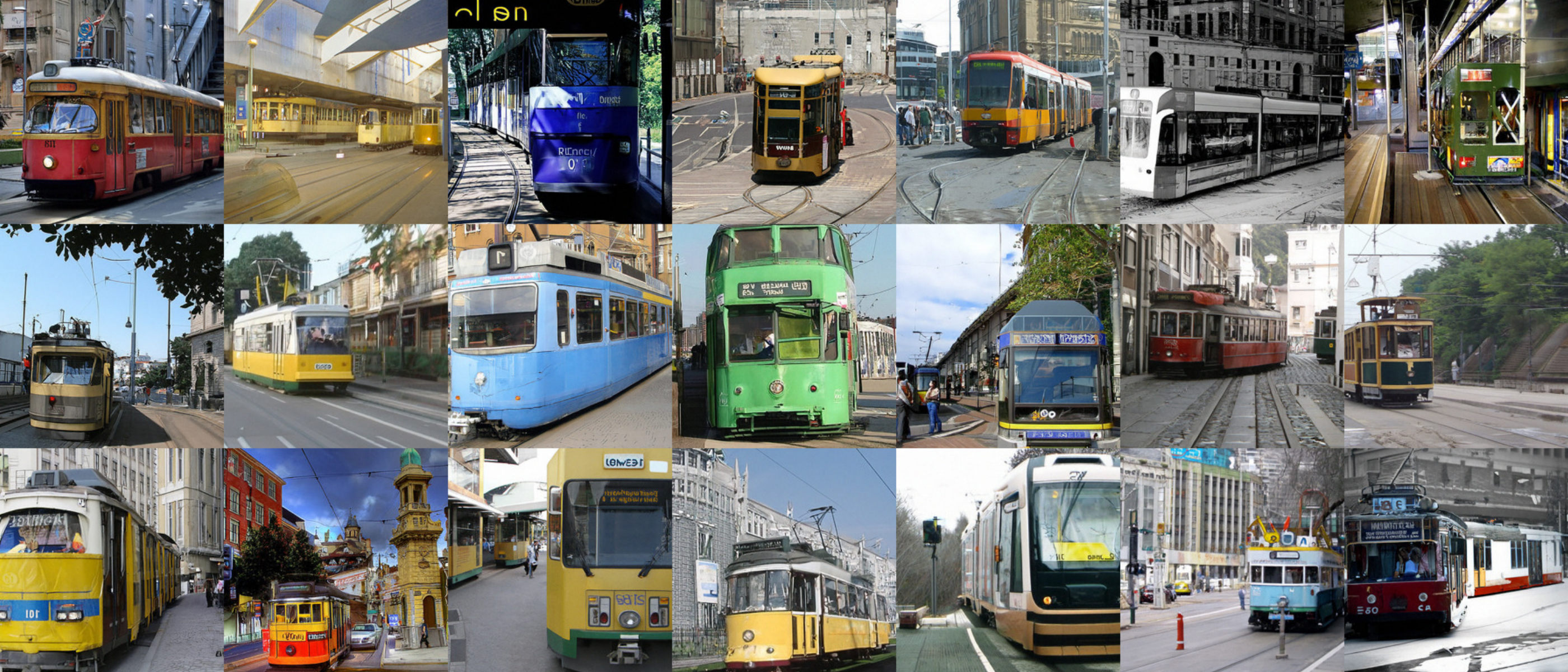}
{829}{Streetcar, Tram, Tramcar, Trolley, Trolley car}
\addclasspdf{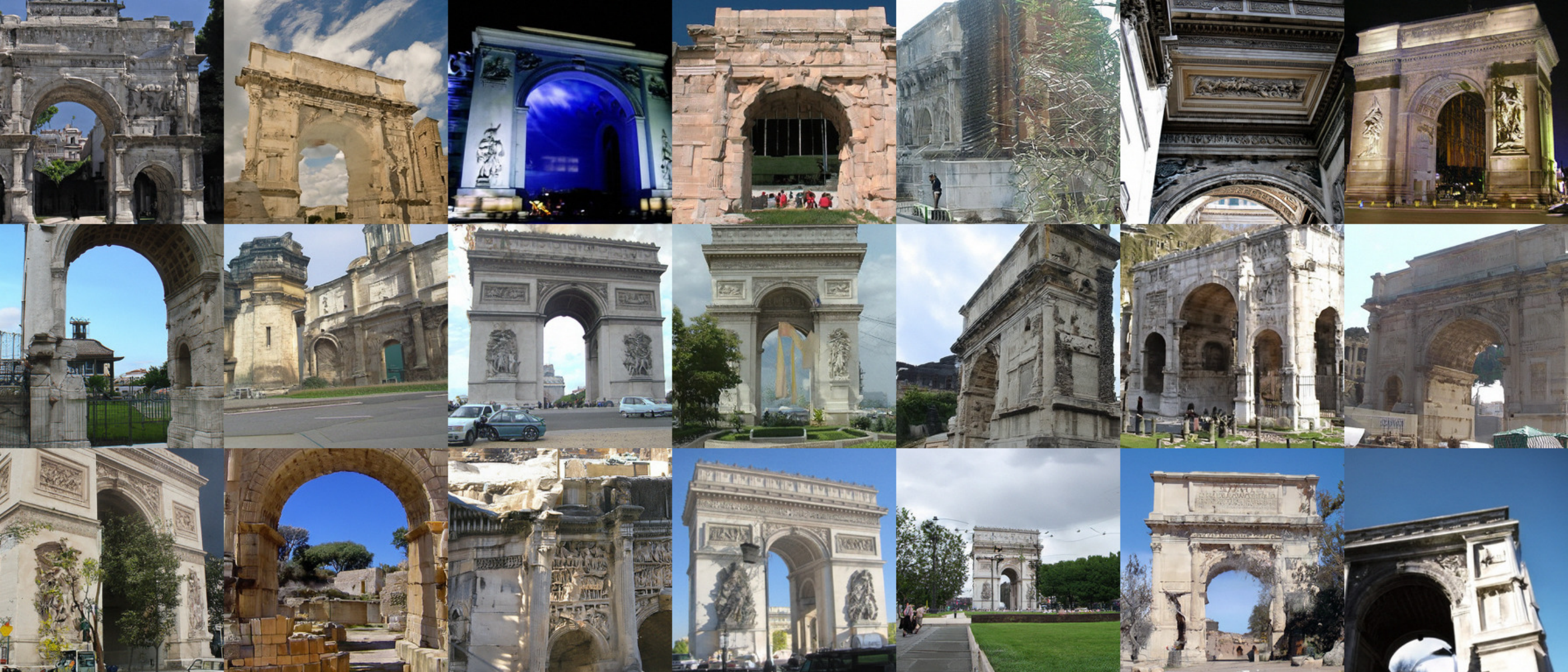}
{873}{Triumphal arch}
\\

\addclasspdf{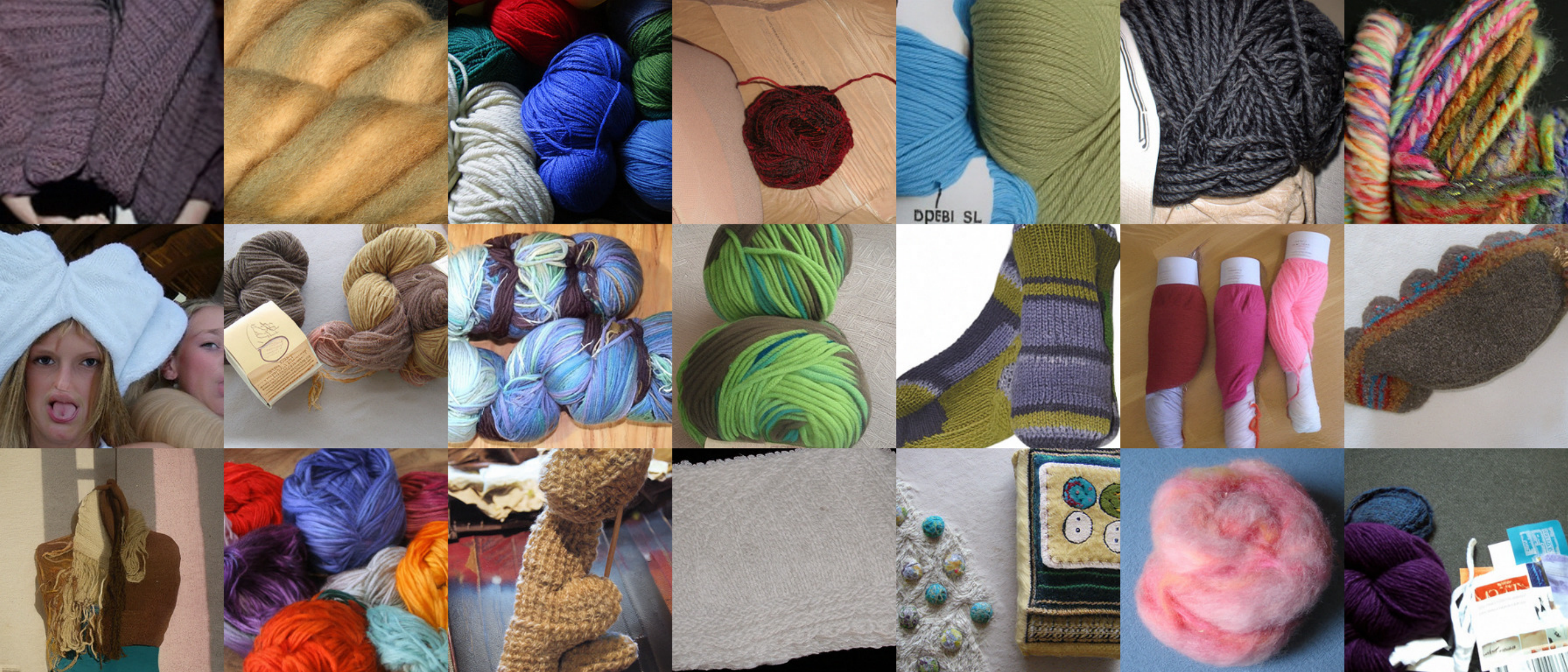}
{911}{Wool, Woolen, Woollen}
\addclasspdf{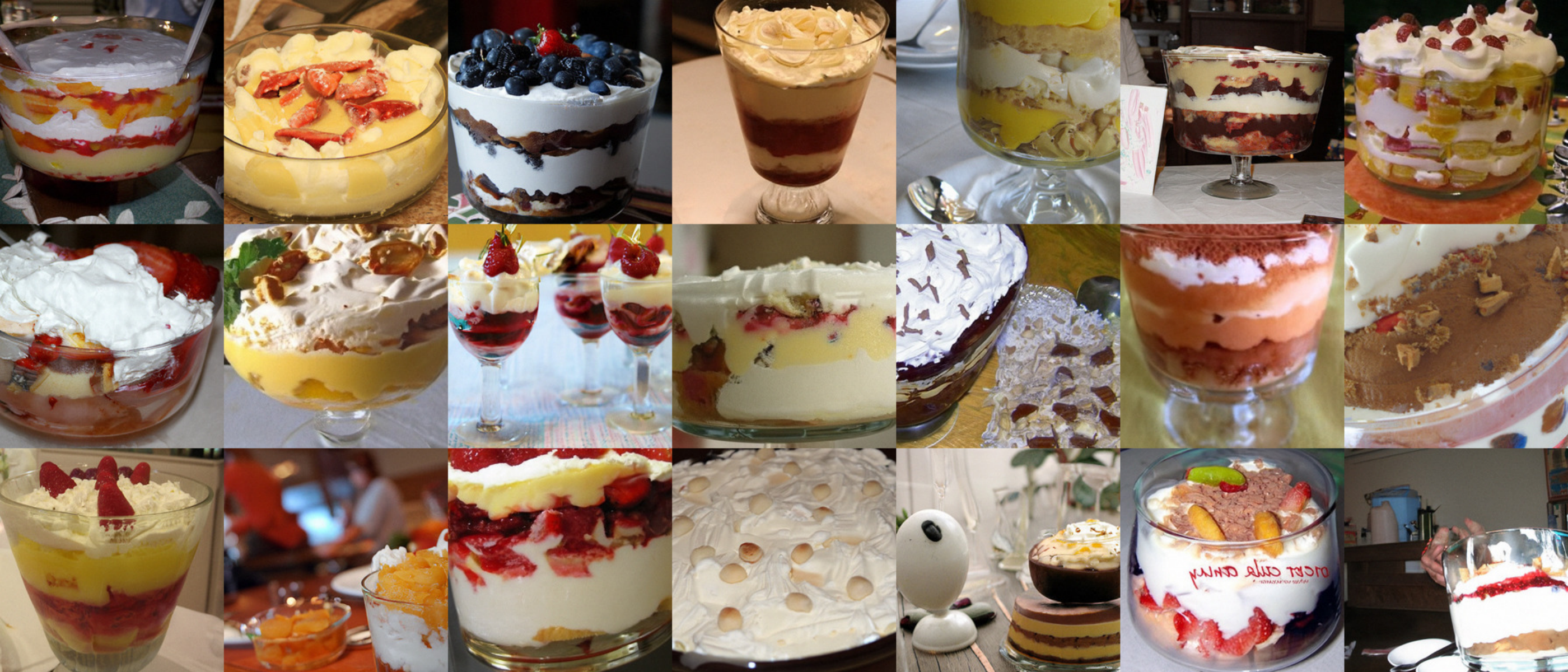}
{927}{Trifle}
\\

\addclasspdf{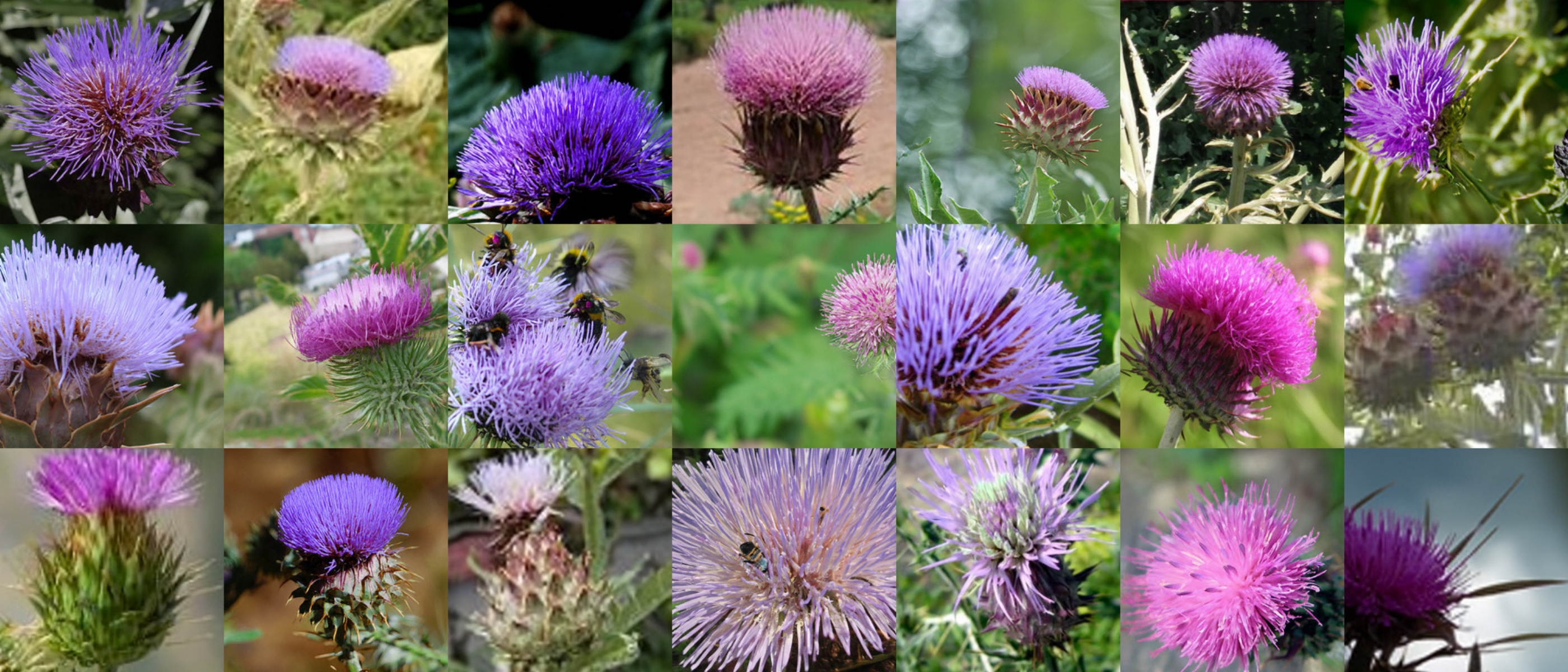}
{946}{Cardoon}
\addclasspdf{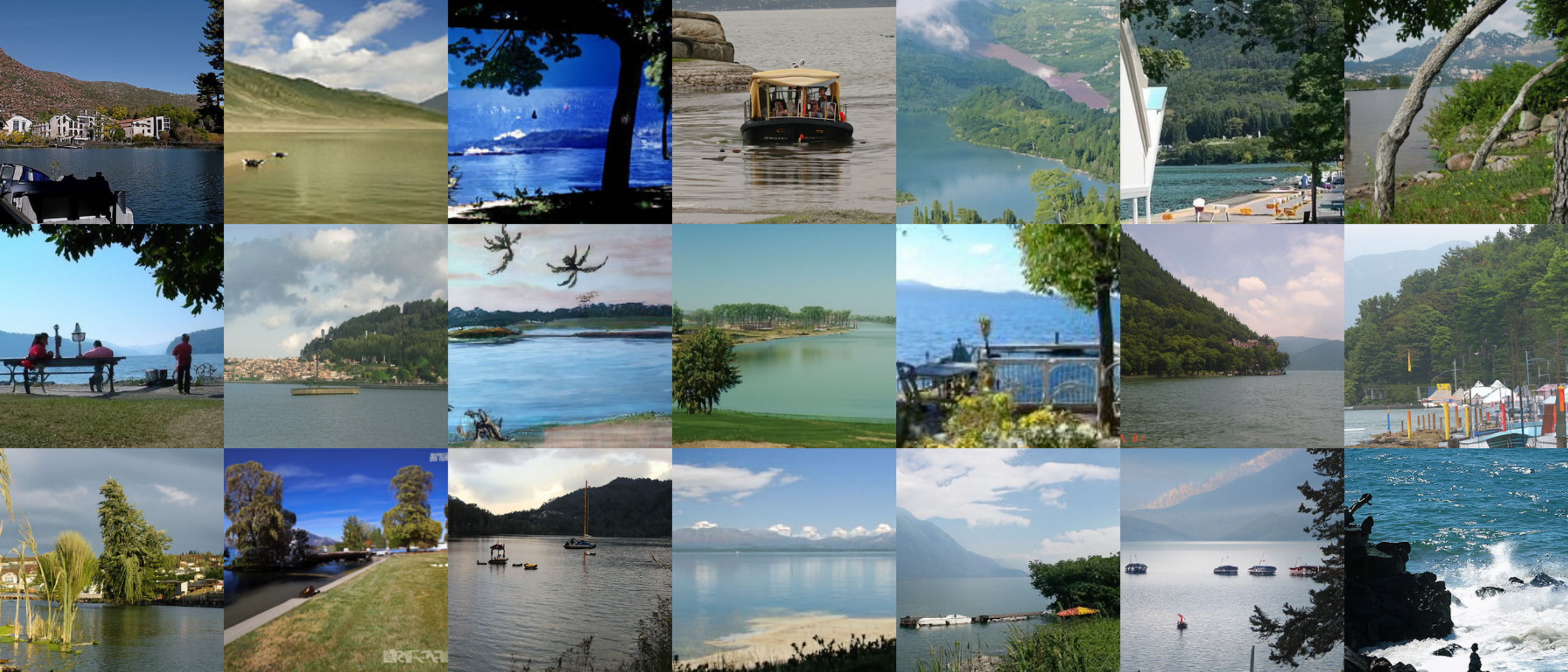}
{975}{Lakeside, Lakeshore}
\\

\caption{
\textbf{\textit{Uncurated} samples of PixelMoE-H-E32A4 on ImageNet-1K at $256\times256$.}
Samples are generated conditioned on the specified classes using the same sampling configuration as the reported quantitative results.
}
\label{fig:more_vis_256_2}
\end{figure*}

\begin{figure*}[p]
\centering

\addclasspdf{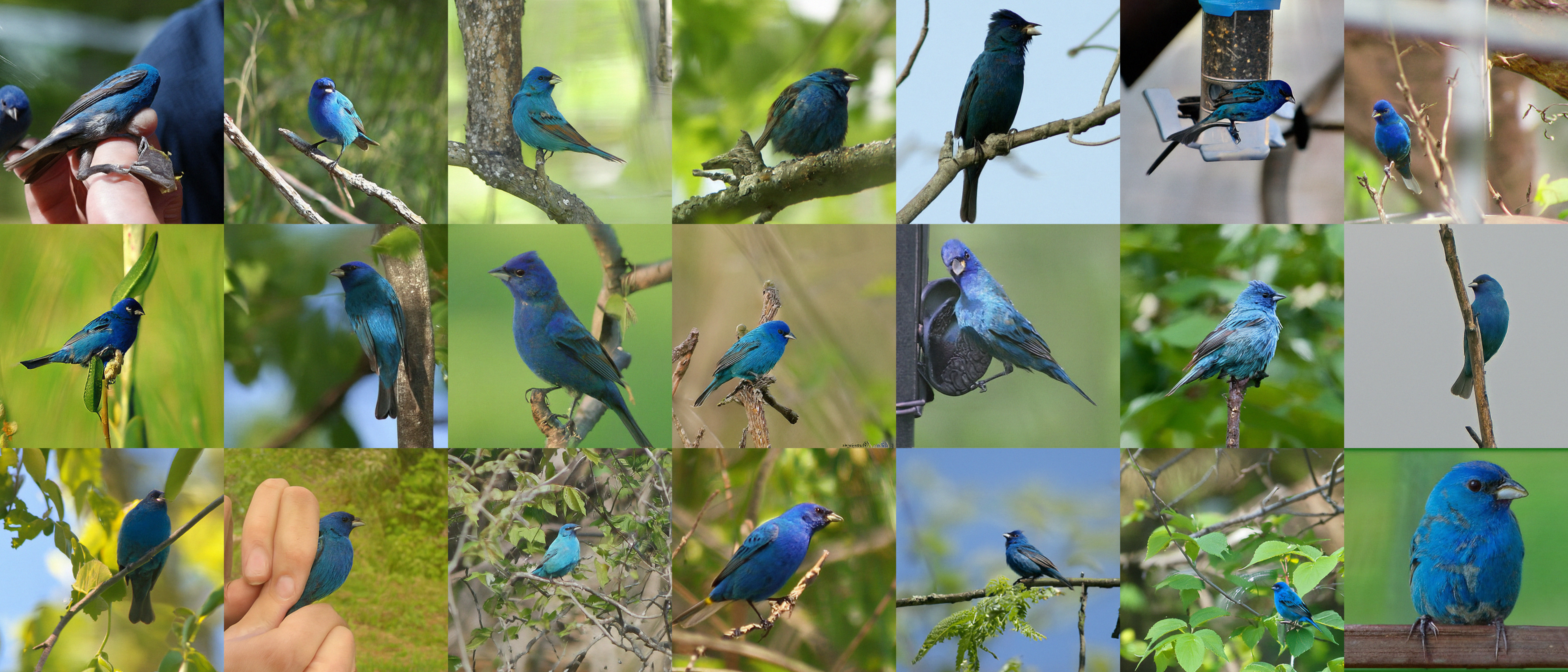}
{014}{Indigo bunting, Indigo finch, Indigo bird, Passerina cyanea}
\addclasspdf{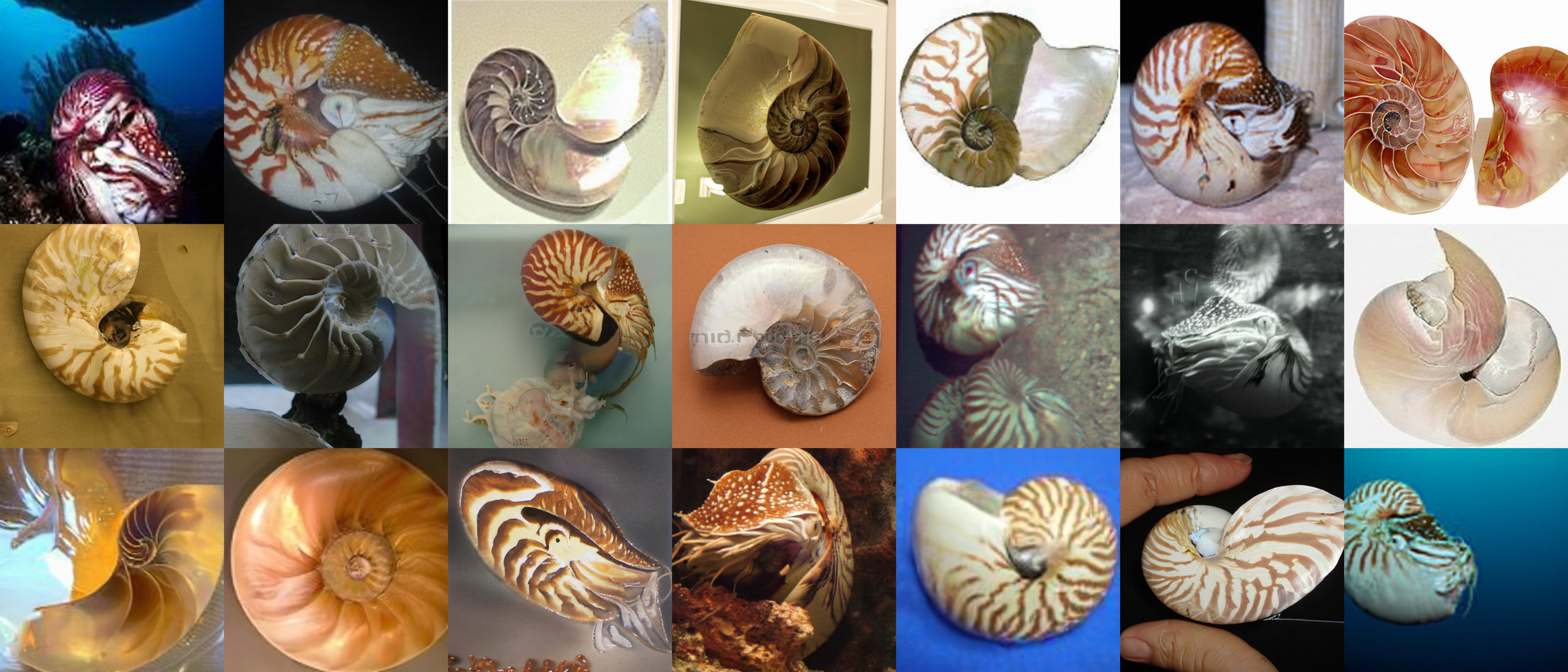}
{117}{Chambered nautilus, Pearly nautilus, Nautilus}
\\
\addclasspdf{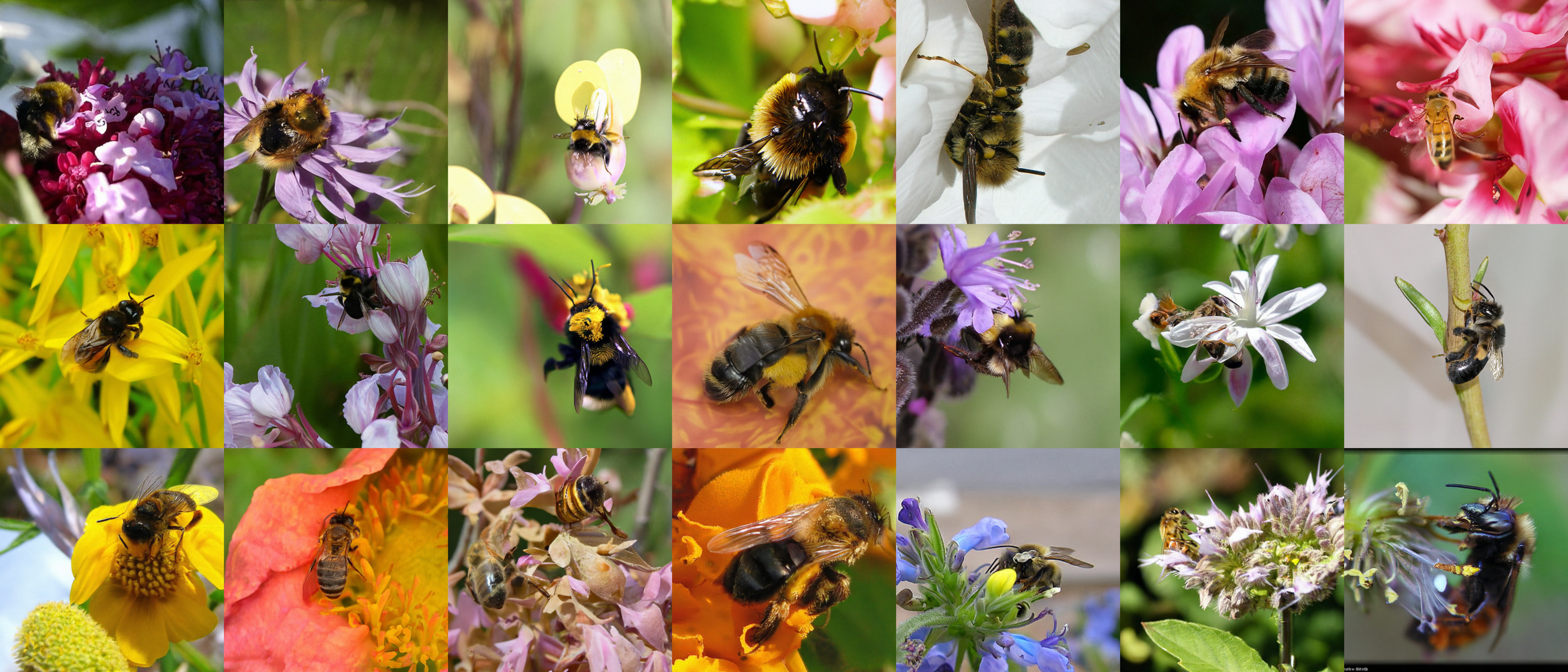}
{309}{Bee}
\addclasspdf{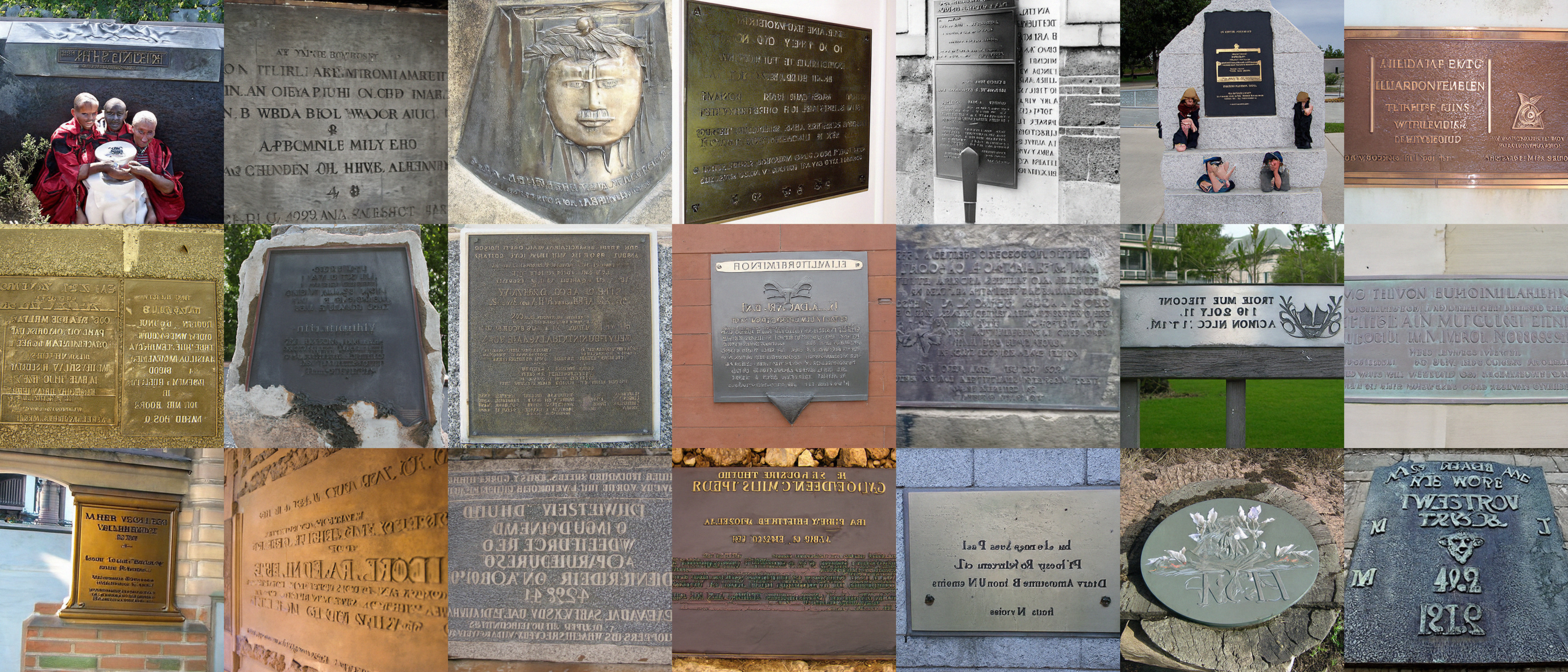}{458}{Brass, Memorial tablet, Plaque}
\\
\addclasspdf{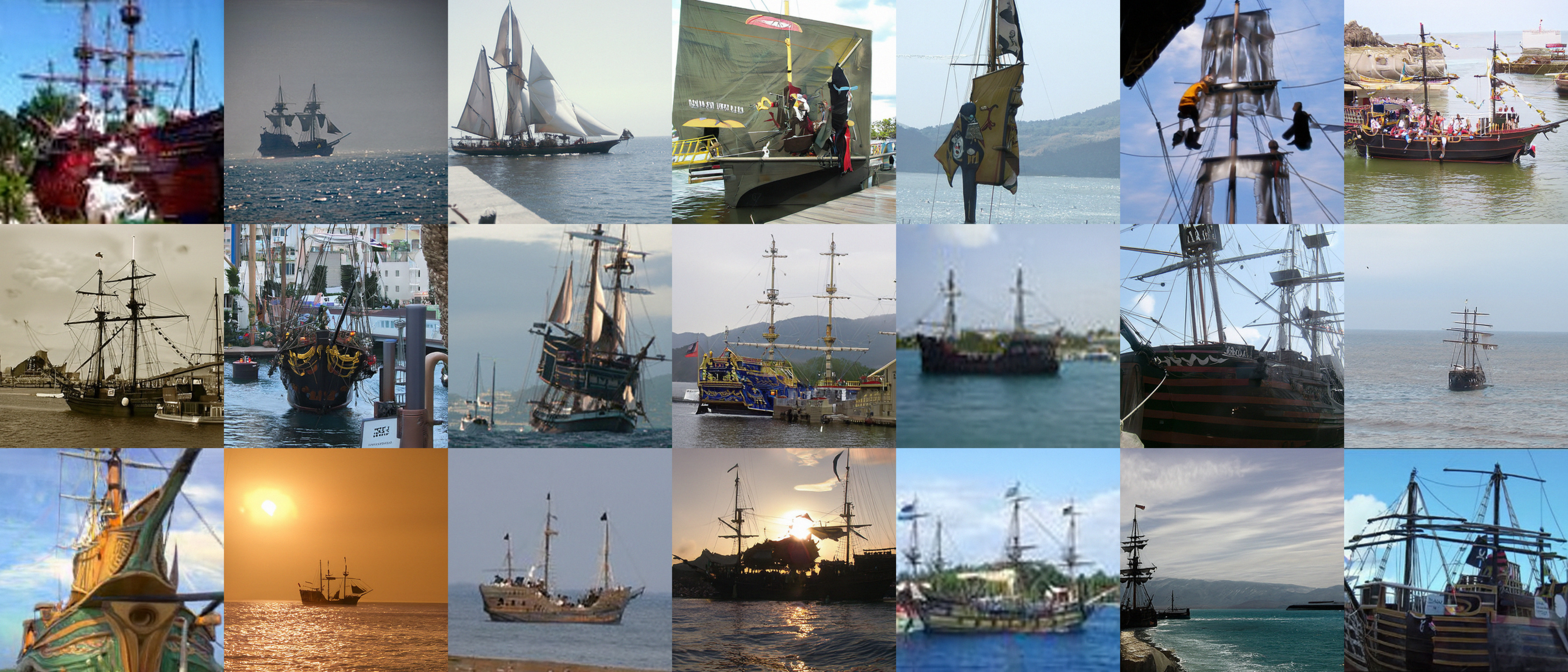}{724}{Pirate, Pirate ship}
\addclasspdf{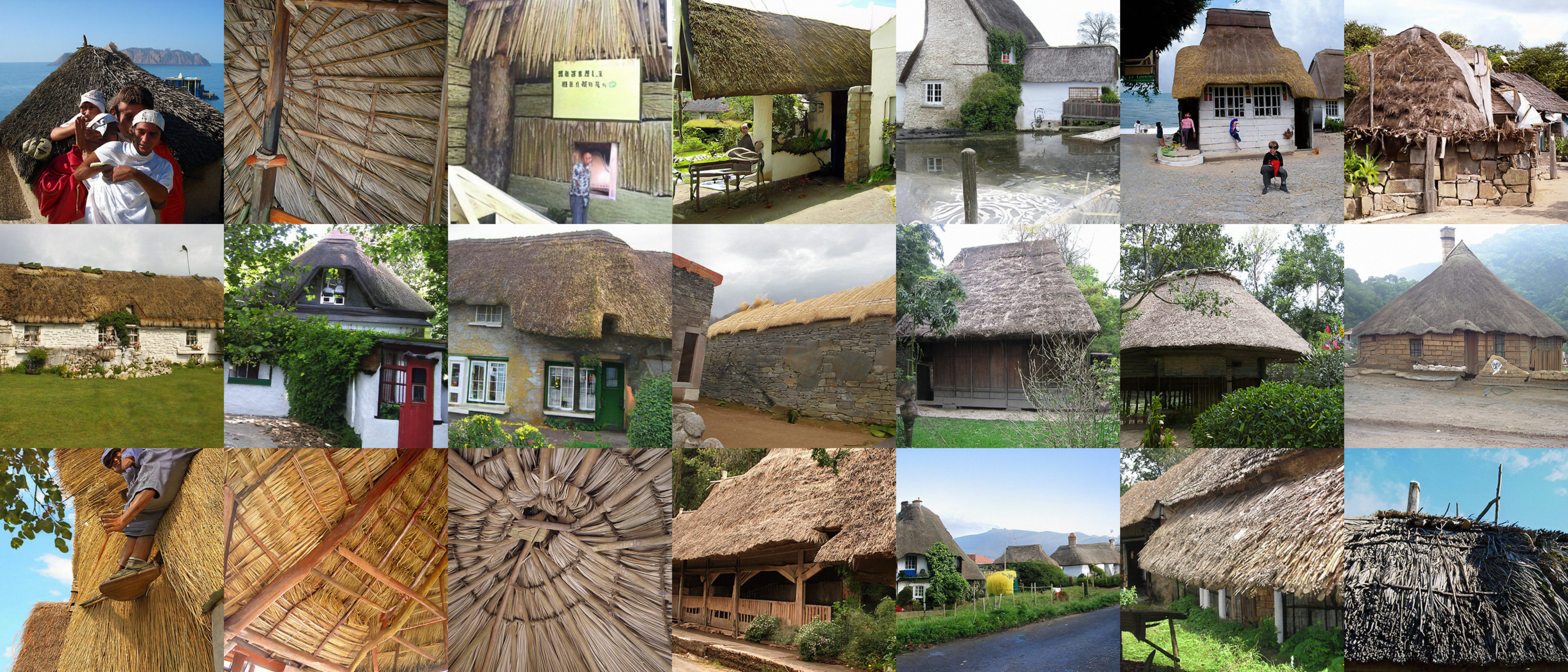}{853}{Thatch, Thatched roof}
\\
\addclasspdf{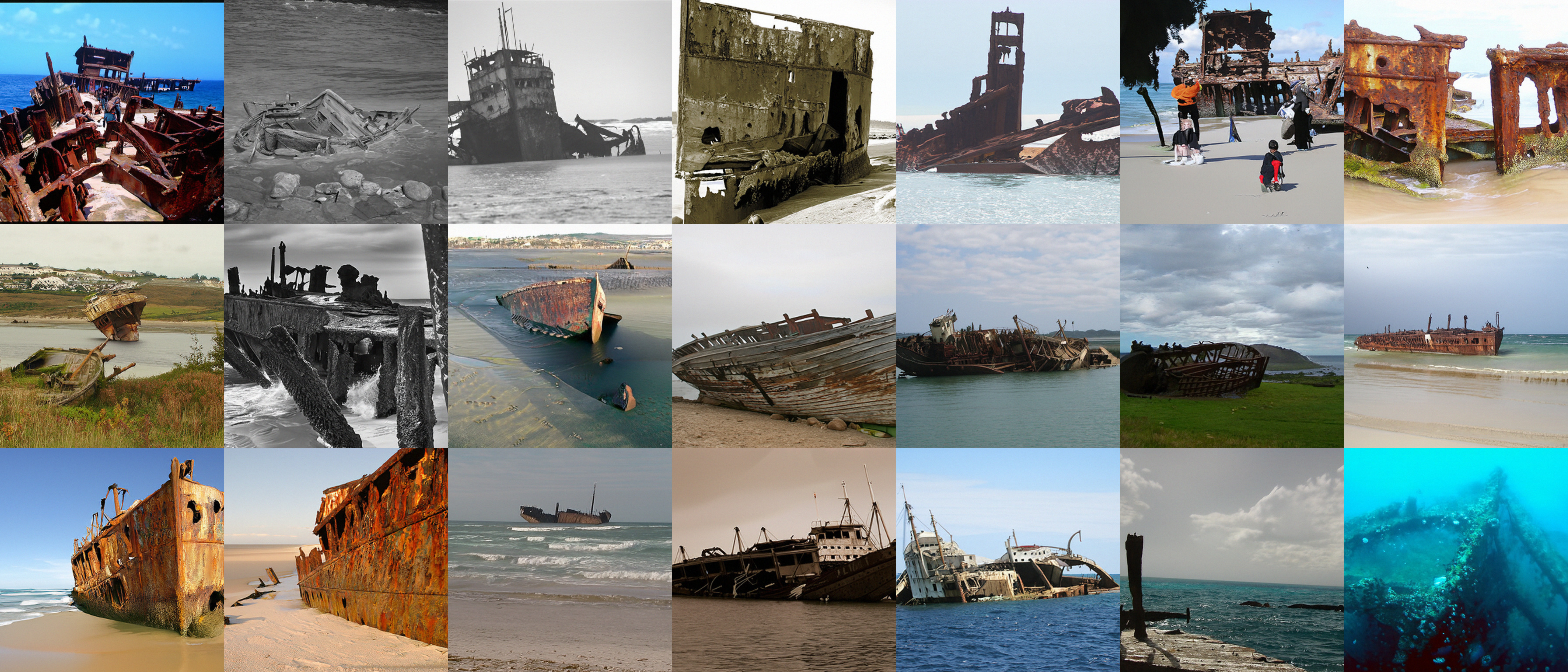}{913}{Wreck}
\addclasspdf{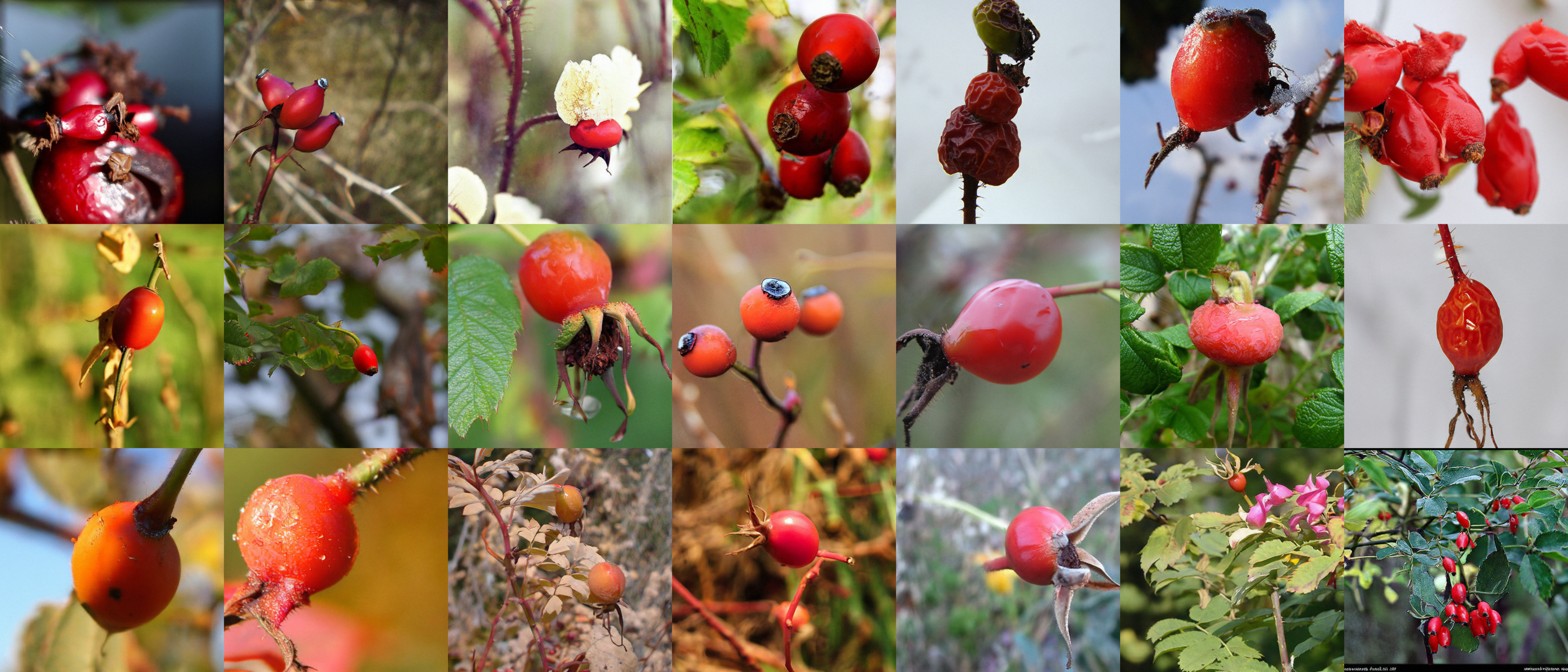}{989}{Hip, Rose hip, Rosehip}
\\

\caption{
\textbf{\textit{Uncurated} samples of PixelMoE-H-E32A4 on ImageNet-1K at $512\times512$.}
Samples are generated conditioned on the specified classes using the same sampling configuration as the reported quantitative results.
}
\label{fig:more_vis_512}
\end{figure*}
\end{document}